\documentclass[preprint,12pt,authoryear]{elsarticle}

\usepackage{amssymb}

\usepackage{subfigure}
\usepackage{graphics}
\usepackage{graphicx}
\usepackage{soul}
\usepackage{enumitem}
\usepackage{multirow}
\usepackage{natbib}
\usepackage{latexsym}
\usepackage{microtype}
\usepackage{array}
\usepackage{colortbl}
\usepackage{soul}
\usepackage{amsmath,amssymb}
\usepackage{graphicx,accents}
\usepackage{hyperref}
\usepackage{dblfloatfix}
\usepackage{tikzpagenodes}
\usepackage{rotating}
\usepackage{comment}

\newcolumntype{P}[1]{>{\centering\arraybackslash}p{#1}}

\newcommand{\name}{\texttt{FLORAL}}

\usepackage{todonotes}
\usepackage{float}
\usepackage{lipsum}
\usepackage{svg}
\usepackage[english]{babel}
\usepackage{amsmath}
\usepackage{graphicx}
\usepackage{hyperref}
\usepackage{booktabs}
\usepackage{xspace}
\usepackage[utf8]{inputenc}
\usepackage[english]{babel}

\usepackage[symbol]{footmisc}

\journal{Elsevier}

\begin{document}

\begin{frontmatter}



\title{See, Hear, Read: Leveraging Multimodality with Guided Attention for Abstractive Text Summarization}

% \newcommand{\symfootnote}[1]{%
% \let\oldthefootnote=\thefootnote%
% \stepcounter{mpfootnote}%
% \addtocounter{footnote}{-1}%
% \renewcommand{\thefootnote}{\fnsymbol{mpfootnote}}%
% \footnote{#1}%
% \let\thefootnote=\oldthefootnote%

% \author[]{Yash Kumar Atri}
% \author[]{Shraman Pramanick}
% \author[]{Tanmoy Chakraborty}
% \author[]{Vikram Goyal}

% \newcommand{\symfootnote}[1]{%
% \let\oldthefootnote=\thefootnote%
% \stepcounter{mpfootnote}%
% \addtocounter{footnote}{-1}%
% \renewcommand{\thefootnote}{\fnsymbol{mpfootnote}}%
% \footnote{#1}%
% \let\thefootnote=\oldthefootnote%
% }
\author{
    Yash Kumar Atri\footnote[1]{Equal contribution; The first author is the contact person.} \quad
	Shraman Pramanick$^*$ \\ 
Vikram Goyal	\quad 
	 Tanmoy Chakraborty  \\
	IIIT-Delhi, New Delhi, India \\
	{\tt \{yashk,shramanp,vikram,tanmoy\}@iiitd.ac.in}
}

% \address{Dept. of Computer Science and Engineering, IIIT-Delhi, India}

\begin{abstract}
In recent years, abstractive text summarization with multimodal inputs has started drawing attention due to its ability to accumulate information from different source modalities and generate a fluent textual summary. However, existing methods use short videos as the visual modality and short summary as the ground-truth, therefore, perform poorly on lengthy videos and long ground-truth summary. Additionally, there exists no benchmark dataset to generalize this task on videos of varying lengths. 

In this paper, we introduce AVIATE, the first large-scale dataset for abstractive text summarization with videos of diverse duration, compiled from presentations in well-known academic conferences like NDSS, ICML, NeurIPS, etc. We use the abstract of corresponding research papers as the reference summaries, which ensure adequate quality and uniformity of the ground-truth. We then propose  {\name}, a factorized multi-modal Transformer based decoder-only language model, which inherently captures the intra-modal and inter-modal dynamics within various input modalities for the text summarization task. {\name} utilizes an increasing number of self-attentions to capture multimodality and performs significantly better than traditional encoder-decoder based networks. Extensive experiments illustrate that {\name} achieves significant improvement over the baselines in both qualitative and quantitative evaluations on the existing How2 dataset for short videos and newly introduced AVIATE dataset for videos with diverse duration, beating the best baseline on the two datasets by $1.39$ and $2.74$ ROUGE-L points respectively.

\end{abstract}


\end{frontmatter}

\section{Introduction}

Abstractive text summarization focuses on generating a summary of a text from its main ideas, not by replicating the most salient sentences, rather by generating new sentences and/or rephrasing existing sentences so that the key semantics and overall meaning of the original text remain intact in summary. It has a wide range of applications in our daily life $-$ from media monitoring, search marketing, email filtering, newsletter production to question-answering chatbots, we are frequently exposed to abstractive summarization. Due to its ability to generate fluent and coherent summaries, abstractive summarization is an important and challenging area of research in {Natural Language Processing} and has got enough attention in the last few years. While most of the previous studies \citep{see2017get, lebanoff-song-liu:2018, gehrmann2018bottom, ijcai2020-0514, CHUNG2020105363} on abstractive summarization use only textual data as the input modality, some studies \citep{shah2016leveraging, li2018read, zhu-etal-2018-msmo, palaskar2019multimodal} have recently focused on incorporating multimodal signals as the input to enhance the quality of text summary. Intuitively, humans can comprehend the gist of an occurrence more quickly by watching relevant images or videos than by only reading a text document, and therefore we believe that multimodal data can also reduce the difficulties for machines to interpret the context.

{\bf Motivation:} With the emergence of multimedia technology and the rapid growth of social media video-sharing platforms such as Youtube and Vimeo, multimedia data (including text, image, audio, and video) have increased dramatically. Specifically, during the COVID-19 outbreak in the last six months, there has been a steep rise in various e-learning platforms, resulting in a drastic increase of online video tutorials and academic presentation videos. However, such videos often do not have the text meta-data associated with them, or the existing ones fail to capture the subtle differences with related videos \citep{wang2012event}. Additionally, different modalities in most of these videos are asynchronous with each other,  leading to the unavailability of subtitles. In this work, we address the task of generating an abstractive text summary of a given academic presentation video so that the viewers can acquire the gist of the presentation in a short time, without watching the video from the beginning to the end. For this purpose, we incorporate automatic speech recognition (ASR) and optical character recognition (OCR) generated text transcripts and capture tonal-specific details of the speaker in addition to extracting semantics and sentics from the video, which are jointly optimized to produce a rich and informative textual summary of the entire presentation. We also show the generalizability of our model on the non-academic dataset (instructional videos). 

{\bf State-of-the-art and Limitations:} The existing studies on abstractive text summarization with multimodal signals include multimodal news summarization \citep{li2018read, li2016multimedia, chen2018abstractive} and summarization of instructional videos \citep{palaskar2019multimodal}. However, all of them use images and/or short videos as the visual modality, which do not generalize on long videos. The generated summaries by these systems are also one or two lines long, and therefore, not suitable for longer academic videos (such as course lecture, conference tutorials). Some other closely related studies include image and video captioning \citep{mun2016text, liu2020sibnet, iashin2020multi, shi2020video, shen2020remote}, video story generation \citep{gella2018dataset}, video title generation \citep{zeng2016generation} and multimodal sentence summarization \citep{li2018multi}; but all of them deal with short videos or images which are not appropriate for our application. The lack of previous studies on this task can be attributed to the absence of a suitable benchmark dataset. In a very recent work, \cite{palaskar2019multimodal} studied the task of summarization of instructional videos on the How2 dataset \citep{sanabria2018how2}, which is the only existing dataset for abstractive text summarization with multimodality. However, the How2 dataset consists of short videos, with an average duration of 90 seconds only. The ground-truth text summaries of this dataset have an average length of 33 words, which are very small as well.

{\bf Our Contributions:} In this paper, we explore the role of multimodality in abstractive text summarization for academic presentation videos of diverse duration and introduce a new resource to further enable research in this area. More specifically, our main contributions in this work are as follows:
\begin{enumerate}[leftmargin=*]
\item We curate the first large-scale dataset, {\bf A}udio {\bf VI}deo l{\bf A}nguage da{\bf T}as{\bf E}t (AVIATE), for abstractive text summarization using multimodal inputs for academic presentation videos of diverse duration. To collect the videos for this dataset, we scraped  $6$ publicly available websites and accumulated paper presentation videos from $28$ well-known international conferences in computer science and social science. To obtain the transcripts of these videos, we apply Deep Speech \citep{DBLP:journals/corr/HannunCCCDEPSSCN14}, a pre-trained end-to-end automatic speech recognition (ASR) system. We use the abstracts of corresponding research papers as the ground-truth summaries, which ensure adequate quality and uniformity. In contrast to How2, AVIATE  contains longer videos and larger ground-truth summaries, which help the deep learning models trained on AVIATE to generalize the performance on other datasets.    
\item We introduce several baselines to show that multimodal frameworks are substantially more effective when compared to their unimodal variants for abstractive text summarization.
\item We propose {\bf Factorized Multimodal Transformer based decoder-only Language Model (\name)}, which uses an increasing number of self-attentions to inherently capture inter-modal and intra-modal dynamics within the asynchronous multimodal input sequences. \name\ demonstrates the utility of pre-trained language model (LM) for summary generation in relatively low-resource setups over traditional encoder-decoder based networks.
\item For the videos of AVIATE, we show the importance of OCR generated text transcripts, which contain keywords and informative phrases displayed on slides in academic presentations. To fuse ASR and OCR generated texts, we introduce a novel {\bf guided attention based fusion mechanism} which attends the complementary features in both the sources and filters out repetitive and redundant words. After the incorporation of OCR transcript, the baselines and \name\ yield $[0.7 - 3.6]$ ROUGE-L points performance improvement on  AVIATE.  
\item \name\ reports benchmark results in terms of both automatic and manual evaluation metrics on How2 for short videos. It beats the best baseline by $1.39$ ROUGE-L points. On AVIATE, \name\ also turns out to be highly effective $-$ it beats the best baseline by $2.74$ ROUGE-L points.
\item Finally, we report the transferability of  \name\   between How2 and AVIATE. When trained on AVIATE and tested on How2, \name\ yields $49.9$ ROUGE-L score, which is only $6.9$ points less than  ROUGE-L obtained when both trained and tested on How2. The diverse-length videos of AVIATE make the model transferable, which is tremendously effective for practical applications.

\end{enumerate}

\if The rest of the paper is organized as follows. Section $2$ summarizes the previous work on abstractive text summarization using both unimodal and multimodal sources. Section $3$ describes the dataset collection, the transcription process, and comparison of AVIATE with How2 on duration and ground-truth summary length statistics. Section $4$ explains features extraction for the different modalities and describes our proposed {\tt FLORAL} network. Section $5$ shows the experimental details and baselines while Section $6$ analyzes it. Finally, Section $7$ offers conclusions and discussion on future research directions. \fi

\noindent
\textbf{Reproducibility:} To reproduce our results, we present detailed hyper-parameter configurations in Table \ref{tab:hyper} and Section \ref{training_params}. Moreover, we also supplement our submission with full AVIATE dataset and source code of \name\footnote{The resources are   available in the following  link: \url{https://github.com/LCS2-IIITD/multimodal_summ}.}.

\section{Related Work}

Abstractive text summarization with multimodal inputs has gained increasing attention in recent years with the surge of multimedia data on the Internet and social media. Unlike unimodal text summarization systems, which are vastly studied, multimodal approaches make use of visual and acoustic modalities in addition to the textual modality, as a valuable source of information for generating a fluent and informative summary. A few existing experiments \citep{li2017multi, li2018read} have shown that compared to unimodal text summarization systems, multimodal summarization can improve the quality of generated summary by using the information in the visual modality.  

\textbf{Unimodal Text Summarization:} Unimodal text summarization systems can be broadly classified into two approaches -- extractive and abstractive summarization. Extractive summarization systems, which are robust and straightforward, involve the selection of phrases and sentences from the source document to generate the summary. Existing literature on extractive summarization has structured the decision either as binary classification over sentences \citep{cheng2016neural, nallapati2016summarunner} or classification followed by ranking \citep{FANG2017189, narayan2018ranking, zhou2018neural, du2020biomedical}. On the other hand, abstractive text summarization systems involve generating novel sentences either by rephrasing or using new words to capture the overall meaning of the source document, making it more advanced and closer to human-like interpretation. \cite{rush2015neural} was the first to apply modern neural networks to abstractive text summarization. Their approach is based on the attention mechanism and has later been augmented with recurrent decoders \citep{chopra2016abstractive}, hierarchical attention networks \citep{nallapati2016abstractive}, variational autoencoders \citep{miao2016language} and pointer-generator (PG) \citep{see2017get} architecture, further improving performance of the summarization task. \cite{song2018structure} proposed a PG-derived structure that tends to preserve structural dependencies from the source into the summaries. \cite{chowdhury2020neural} improved the work of \cite{song2018structure} by adding a structural attention based encoder to implicitly capture long term dependency relations in the summarization of lengthy documents. Recently, transformers \citep{vaswani2017attention} have been used to effectively encode sequential data with great success when pre-trained for language modeling or language masking and subsequently fine-tuned \citep{radford2018improving, devlin2018bert} and thus, can be used on relatively low-resource setup without overfitting.

\textbf{Text Summarization with Multimodality:} Abstractive text summarization with multimodality deals with the fusion of textual, acoustic and visual modalities for summarizing a video document with a text precise that outlines the content of the entire video. Multimodal information is very useful in learning human-like meaning representations \citep{baroni2016grounding, kiela2017deep, pramanick2021exercise, pramanick-etal-2021-detecting}. Since text rarely occurs in isolation in the real world, it becomes very effective to use all available information to optimize the quality of the summary jointly. The existing literature on multimodal text summarization include multimodal news summarization \citep{li2018read, chen2018abstractive, li2016multimedia} and summarization of instructional videos  \citep{palaskar2019multimodal}. \cite{li2017multi} were the first to collect a multimodal corpus of 500 English news videos and articles with manually annotated the summaries. However, the size of this dataset is very small. \cite{zhou2017towards} presented the YouCookII dataset, containing instructional videos for cooking recipes with temporarily localized annotations for the procedures. \cite{zhu-etal-2018-msmo} introduced the notion of multimodal summarization with multimodal output, which takes the news with images as input, and finally outputs a pictorial summary. Most recently, \cite{palaskar2019multimodal} studied the task of summarization of instructional videos on the How2 dataset \citep{sanabria2018how2}, which can be considered as the closest task to ours. However, all existing multimodal text summarization methods focus on summarizing images and/or short videos and generate one- or two-line long summary, and thus, can not be generalized to longer videos.

\begin{figure*}[!t]
  \centering
  \vspace{-5mm}
  \scalebox{0.83}{
  \subfigure[Duration statistics across the dataset.]{\includegraphics[scale=0.39]{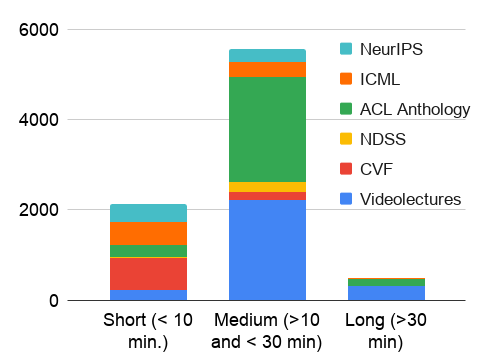}}\hspace{0.3cm}}
  \scalebox{0.83}{
  \subfigure[Source statistics across the dataset.]{\includegraphics[scale=0.33]{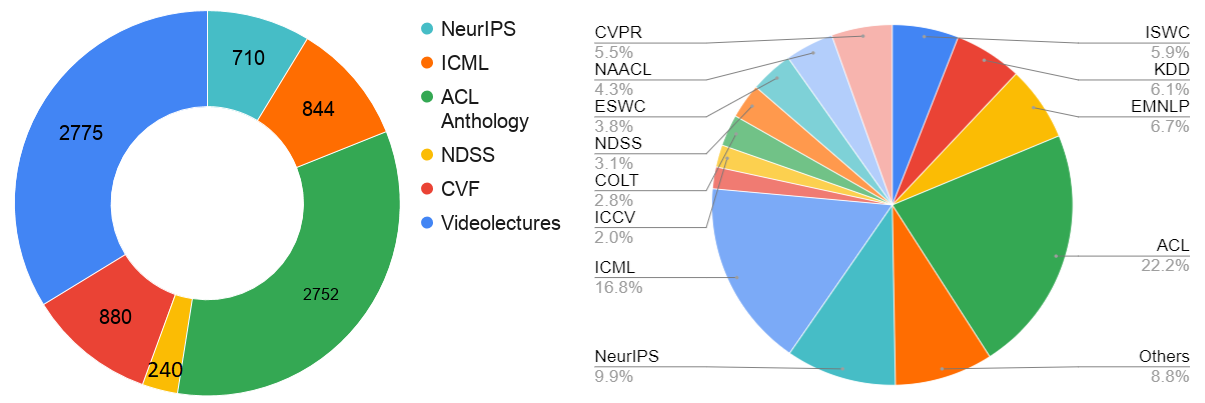}}}
  \vspace{-3mm}
  \caption{Duration and source statistics of AVIATE.}
  \label{figure:dataset}
  \vspace{-2mm}
\end{figure*}

%\vspace{-1mm}
\section{Dataset}

To enable the exploration of abstractive text summarization using multimodal signals and to generalize the task for videos of different lengths, we introduce AVIATE, the first large-scale multimodal text summarization dataset with videos of diverse duration, compiled from academic paper presentations. Currently, the only existing benchmark dataset relevant to our task is the How2 dataset \citep{sanabria2018how2}, which includes short instructional videos on different topics like cooking, sports, indoor/outdoor activities, music, etc. Our study reveals that deep neural models trained on such short videos fail to produce satisfactory results on longer videos. Moreover, the facial expression of the speakers in academic talks and presentations often plays an important role to preserve the most informative frames, which is not always the case for the How2 dataset.

%\vspace{-2mm}
\subsection{How2 Dataset}

The How2 dataset consists of $2,000$ hours of short instructional videos, where the training, validation, and test set contain $73,993$, $2,965$, and $2,156$ videos respectively, with an average length of $90$ seconds. Each video in this dataset is accompanied by a human-generated transcript and a $2-3$ sentence ground-truth summary. The average length of transcripts and summaries is $291$ and $33$ words respectively.
%\vspace{-2mm}
\subsection{AVIATE Dataset}\label{sec:avation}
To collect conference presentation videos, we identified $28$ academic conferences in computer science and social science, spanning over various domains such as Machine Learning, Natural Language Processing, Data Mining, Computer Vision, Computational Linguistics, Semantic Web, and Complex Systems. Most of the videos of our dataset come from conferences like NDSS, ICML, NeurIPS, ACL, NAACL, CVPR, EMNLP, ISWC, KDD, etc. To collect the oral and spotlight presentations of these conferences, we scrapped six different academic online video repositories, namely Videolectures.NET\footnote{\url{http://videolectures.net/}}, ACL Anthology\footnote{\url{https://www.aclweb.org/anthology/}}, CVF Open Access\footnote{\url{https://openaccess.thecvf.com/}}, ICML\footnote{\url{https://icml.cc/}}, NeurIPS\footnote{\url{https://nips.cc/}}, and NDSS Symposium\footnote{\url{https://www.ndss-symposium.org/}} websites. All the paper presentation videos are accompanied by an abstract, which we use as the ground-truth summary. Thus, unlike the How2 dataset, we did not annotate the summaries ourselves, which significantly improved the quality of ground truth summaries, and hence of the entire dataset. 

AVIATE consists of a total of $8,201$ videos, which spans over almost $2,300$ hours. Among them, we use $6,680$ videos for training, $662$ for validation, and $859$ for testing. The length of summaries is mostly between $100-300$ words, with an average of $168$ words. A brief source and duration statistics of AVIATE is presented in Figure \ref{figure:dataset}.\\

\noindent\textbf{Transcription:} 
%A few videos in the dataset have an available transcript. Therefore, we use automatic speech recognition (ASR) to generate text transcript
Since we collected all the videos from six different sources, not all of them had subtitles or transcripts readily available. This is particularly the case for videos from ACL Anthology and Videolectures, which contribute the majority of the AVIATE dataset. In the case of videos from NDSS, ICML, NeurIPS, and CVF Open Access corpus, subtitles are available for those videos which are present on Youtube. To maintain uniformity in the quality of transcripts, we apply Deep Speech \citep{DBLP:journals/corr/HannunCCCDEPSSCN14}, a pre-trained end-to-end speech recognition algorithm, to extract transcripts for all the videos. To ensure the quality of Deep Speech generated transcripts, we manually transcribe $300$ randomly selected videos from our dataset. A low word error rate ($24.19\%$) of the Deep Speech model for those videos indicates the satisfactory standard of the transcripts. An additional normalization step, which includes formatting\footnote{For example, labeling `September 16, 2017' as `september sixteenth twenty seventeen'.} entities like numbers, dates, times, and addresses, helps us to further reduce the error rate to $20.12\%$.
%, which is perfectly admirably sufficient for our application.  

\begin{figure}[!t]
\centering
\scalebox{0.9}{
\subfigure[Video length and ground-truth summary length distribution for AVIATE.]{\includegraphics[width =\columnwidth]{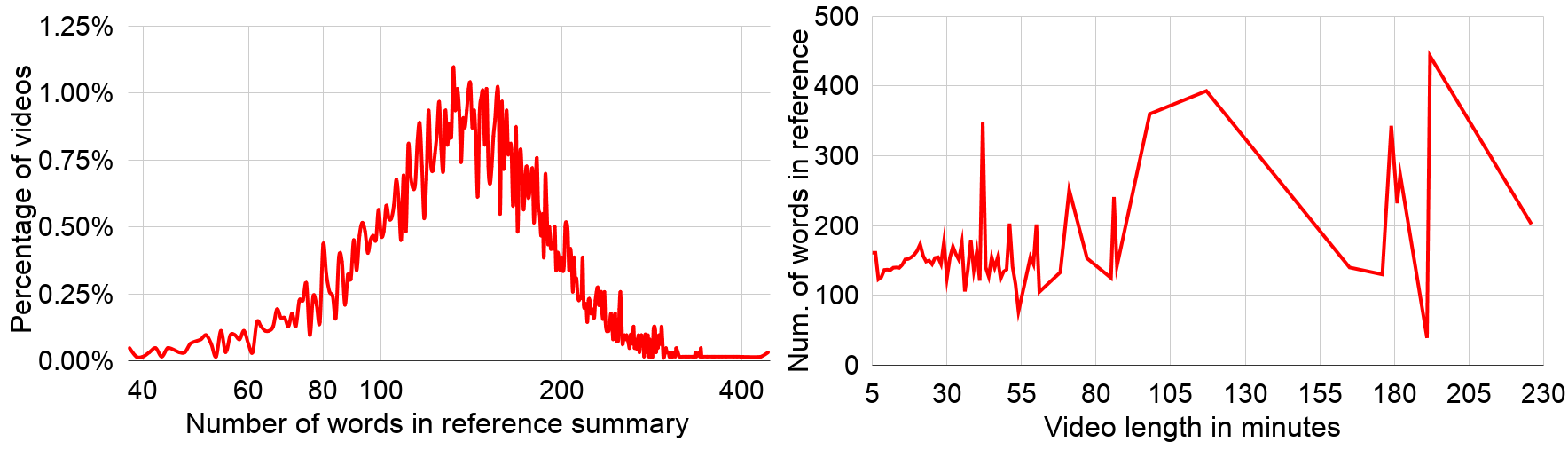}}}\hspace{1.8em}
\scalebox{0.9}{
\subfigure[Video length and ground-truth summary length distribution for How2.]{\includegraphics[width =\columnwidth]{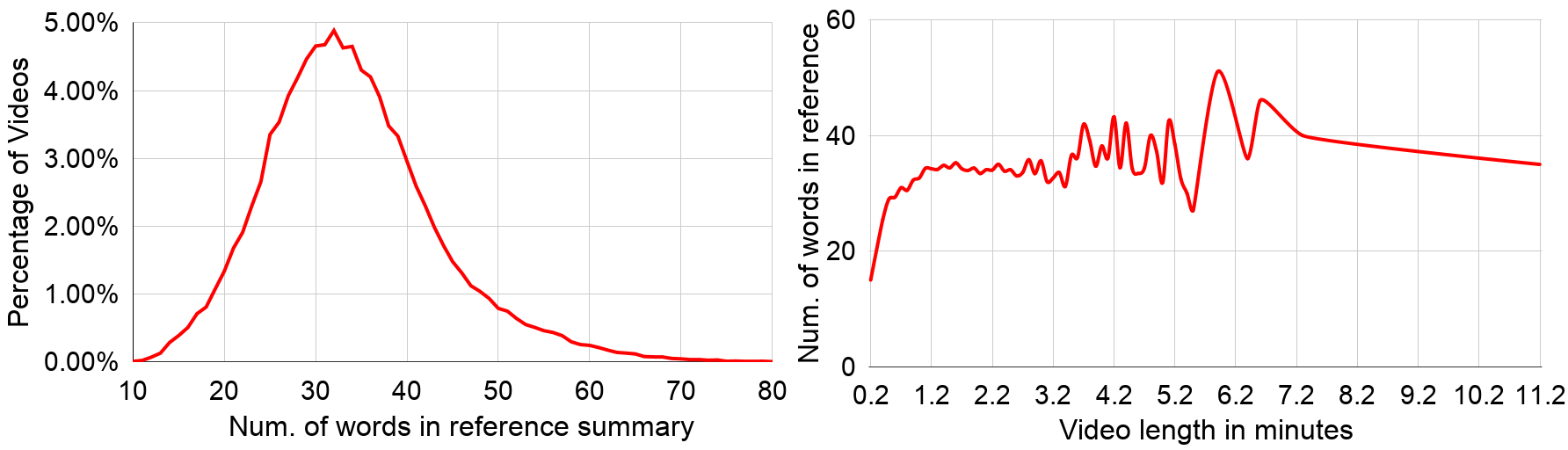}}}
\vspace{-2mm}
\caption{Correlation between duration of videos and ground-truth summary length for AVIATE and How2 datasets.}
\label{figure:dataset_summary}
\vspace{-2mm}
\end{figure}

Figure \ref{figure:dataset_summary} shows a comparison of video length and ground-truth summary length distribution for AVIATE and How2. For both datasets, longer videos generally have a longer ground-truth summary, which leads to an overall positive correlation between video duration and ground-truth summary length. The average length of AVIATE videos is almost $12$ times more than that of How2 videos. The longer videos and lengthier summaries in AVIATE make it harder than How2 to train on, which is explained in Section \ref{section:results}. 

\section{\textbf{\name}: Our Proposed System}

\begin{figure*}[hbtp]
  
  \centering
  \hspace{-3.5mm}
  \scalebox{0.98}{
  \subfigure[\scriptsize{Overview of our proposed model, \name.}]
  {\includegraphics[width =0.60\textwidth]{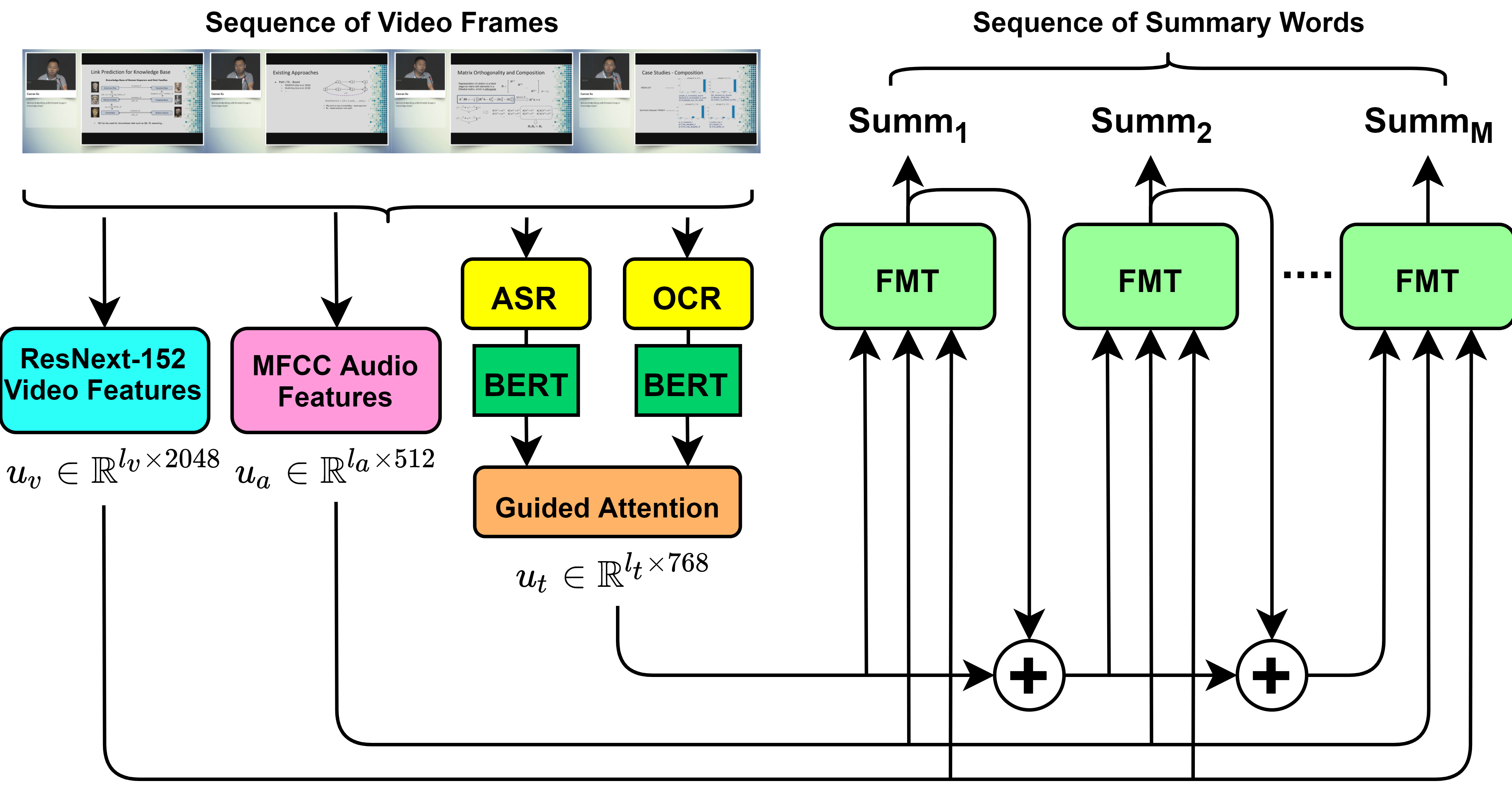}}\hspace{0.16cm}\label{fig:model}}
  \scalebox{0.955}{
  \subfigure[\scriptsize{Architecture of MTL.}]{\includegraphics[width =0.19\textwidth]{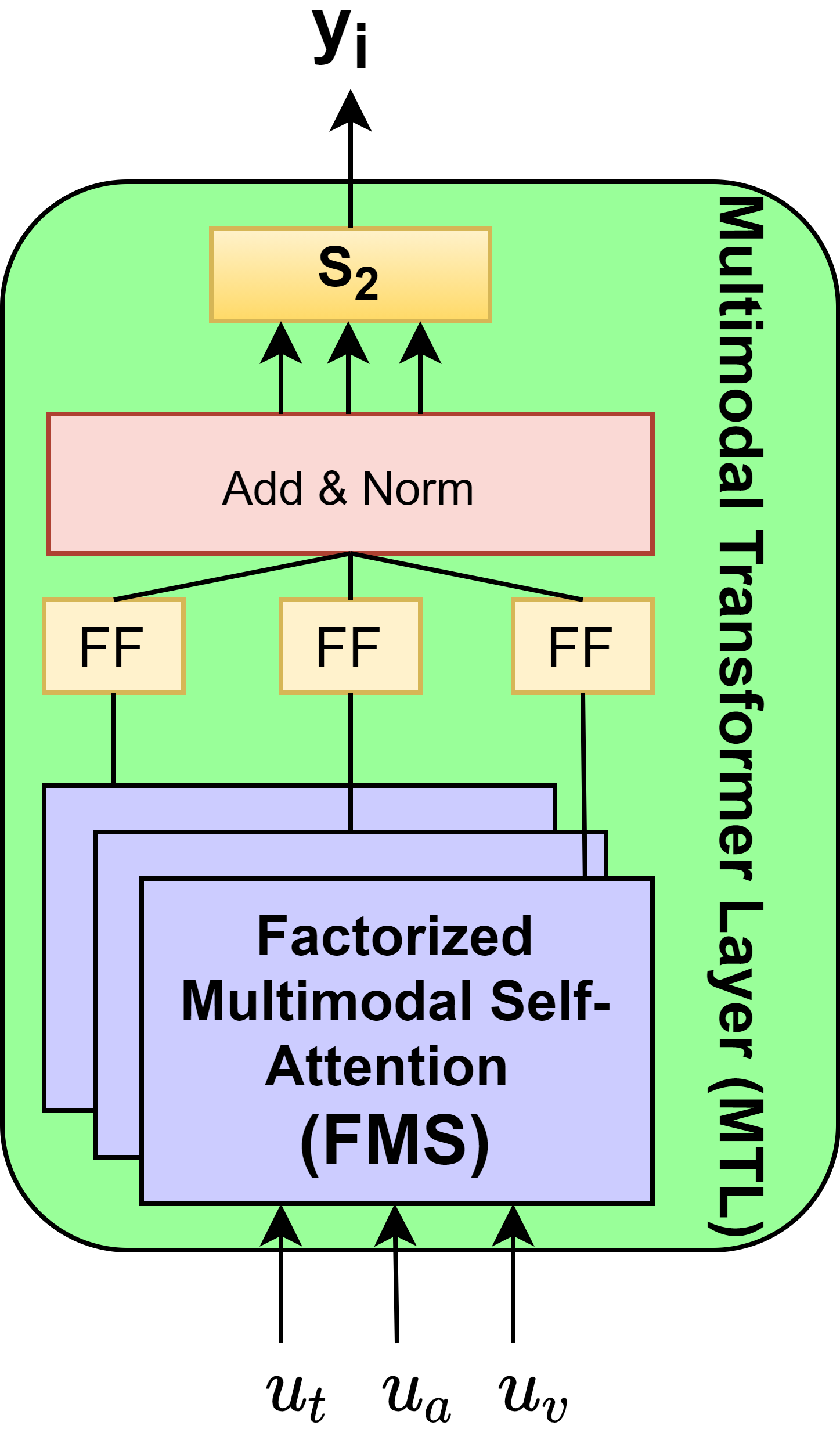}\label{fig:mtl}}\hspace{0.16cm}}
  \scalebox{0.955}{
  \subfigure[\scriptsize{Architecture of guided attention.}]{\includegraphics[width =0.18\textwidth]{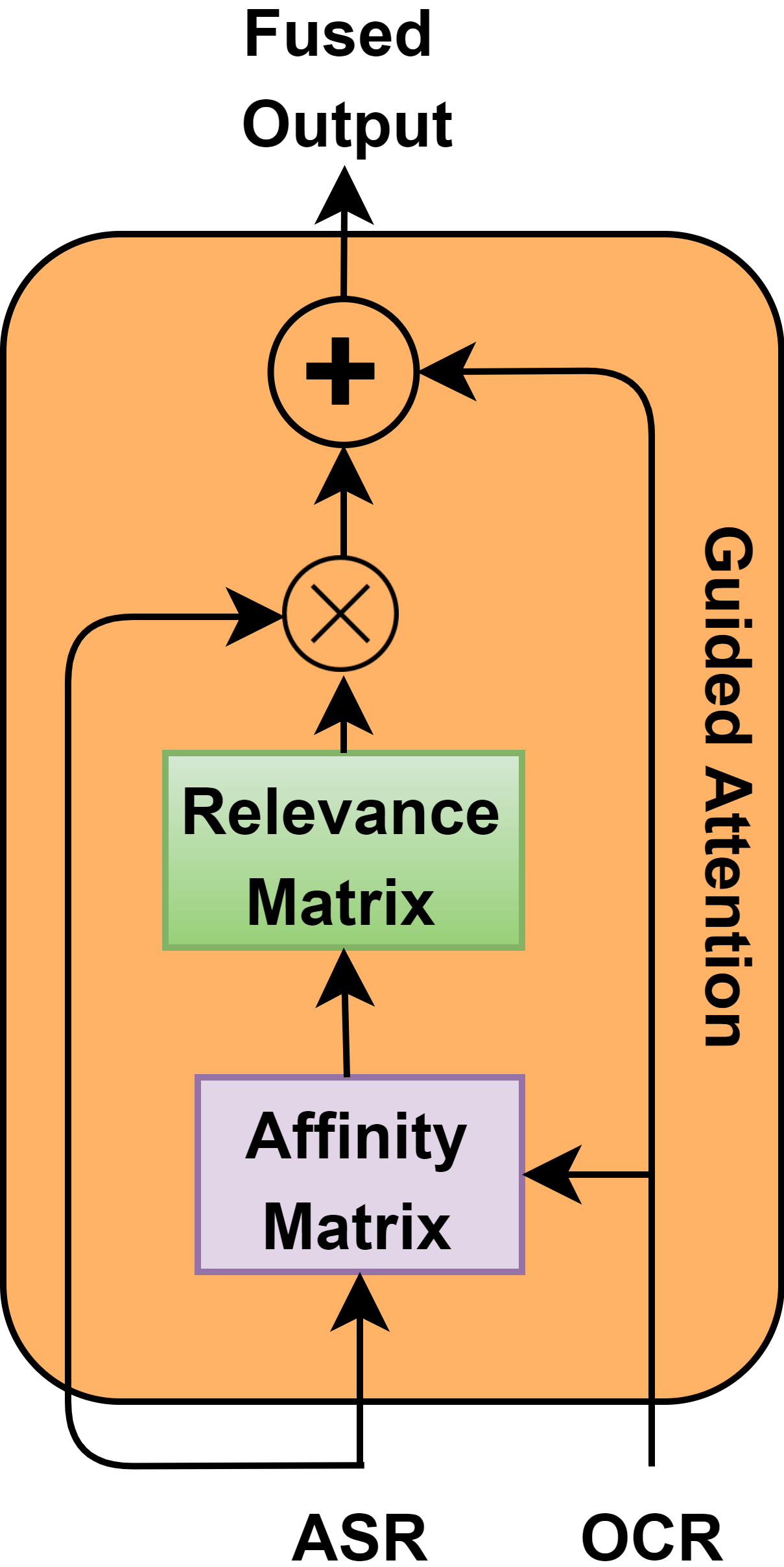}\label{guided}}}
  \vspace{-2mm}
  \caption{The complete architecture of \name, our proposed Factorized Multimodal Transformer based decoder-only Language Model. The Factorized Multimodal Transformer \citep{zadeh2019factorized} consists of a stack of Multimodal Transformer Layers (MTL), which is shown in Figure \ref{fig:mtl}. Figure \ref{guided} shows the architecture for guided attention layer used for the fusion of ASR and OCR generated text transcripts.} 
  \vspace{-2mm}
  \label{figure:model_total}
\end{figure*}

% \begin{figure*}[hbtp]
%   \centering
%   \hspace{-10mm}
%   \scalebox{1.2}{
%   \subfigure[\footnotesize{Overview of our proposed model, \name.}]
%   {\includegraphics[width =1\textwidth]{table_figures/model.png}}\hspace{0.25cm}\label{fig:model}}
%   \scalebox{1.2}{
%   \subfigure[\footnotesize{Architecture of MTL.}]{\includegraphics[width =1\textwidth]{table_figures/mtl_last.png}\label{fig:mtl}}\hspace{0.25cm}}
%   \scalebox{1}{
%   \subfigure[\footnotesize{Architecture of \\ guided attention.}]{\includegraphics[width =1\textwidth]{table_figures/guided_last.png}\label{guided}}
%   \vspace{-4mm}}
%   \caption{The complete architecture \name, our proposed Factorized Multimodal Transformer based decoder-only language model. The Factorized Multimodal Transformer \citep{zadeh2019factorized} consists of a stack of Multimodal Transformer Layers (MTL), which is shown in Figure \ref{fig:mtl}. Figure \ref{guided} shows the architecture for guided attention layer used for the fusion of ASR and OCR generated text transcripts.} 
%   \vspace{-2mm}
%   \label{figure:model_total}
% \end{figure*}

In this section, we describe our proposed system, {\bf Factorized Multimodal Transformer \citep{zadeh2019factorized} based Language Model (\name)} for abstractive text summarization using multimodal signals. Figure \ref{figure:model_total} shows the overall architecture of \name. {It takes a video, its corresponding audio and text transcript as input and generates an abstractive textual summary.} A video generally has three distinct modalities -- visual, textual, and acoustic, which supplement each other by providing complementary information, and thus when fused, separately contribute to generating richer and more fluent summaries. The first part of \name\ extracts unimodal features using respective unimodal feature extraction networks. This phase does not consider the contextual relationship between the three different modalities. In the next part, unimodal features are processed using the Factorized Multimodal Transformer ({\tt FMT}) based decoder-only network over multiple steps, which in turn generates one summary word in each step. After every step, the generated word is appended to the source text with a delimiter. Therefore, \name\ considers the entire summarization problem as a language modeling task, simplifying traditional encoder-decoder architecture. The remaining part of this section discusses individual modules of \name\ in detail.  

\subsection{Video Feature Extraction}
The visual features are extracted using a pre-trained action recognition model, $ResNeXt-152$ $3D$ Convolutional Neural Network \citep{hara3dcnns} trained on the Kinetics dataset \citep{kay2017kinetics} to recognize $400$ different human actions. All the frames, computed at a rate of 5 FPS, are first preprocessed by resizing, center-cropping, and normalization to have a resolution of $112 \times 112$. For every 16 non-overlapping frames in a video, $ResNeXt-152$ extracts a 2048 dimensional feature vector. Therefore, the result is a sequence of feature vectors per video rather than a global one. The sequential feature vector, $u_v = \{u_i^v\}_{i = 1}^{l_v}$, is then used as the visual embedding input to the FMT.  

\subsection{Speech Feature Extraction}
The acoustic modality is expected to contribute information related to tonal-specific details of the speaker \citep{tepperman2006yeah}. To achieve this, we obtain low-level features from the audio stream for each video. Similar to \cite{castro2019towards}, we use the popular speech processing library, Librosa \citep{mcfee2018wzy} and perform the steps mentioned next. First, the audio sample for a video is stacked as a time-series signal with a sampling rate of $16000$ Hz. Next, we remove the echos and background noise from the audio signals by integrating it with Audacity instance\footnote{\url{https://github.com/officeonlinesystems/audacityonline_audioeditor/}}, which is a free and open-source audio editor. Then, we segment the audio signals into $d_w$ non-overlapping windows with a window size of $25$ ms and successive window shift of $10$ ms to extract low-level features that include Mel Frequency Cepstral Coefficients (MFCCs) with hamming window and the related temporal derivatives. Padding and segmentation are performed to achieve a fixed-length representation of the audio sources which are otherwise variable in length. At last, we concatenate all the extracted features to compose a $d_a = 512$ dimensional joint representation for each window. Final MFCC features are obtained by applying a log Mel frequency filter bank over $0$ to $8000$ Hz and applying discrete cosine transformation (DCT). Similar to the visual features, the audio features,  $u_a = \{u_i^a\}_{i = 1}^{l_a}$, are also sequential for every video sample and are then used as the acoustic embedding input of FMT.

\subsection{Textual Feature Extraction}
Both How2 and AVIATE datasets contain textual transcripts corresponding to video samples. For How2, the transcripts are manually annotated, while for AVIATE, a pre-trained automatic speech recognition (ASR) algorithm, Deep Speech \citep{DBLP:journals/corr/HannunCCCDEPSSCN14}, is used to extract the transcripts for all the videos (as discussed in Section \ref{sec:avation}). Since the AVIATE dataset consists of conference presentation videos, we observe that in the majority of video samples of AVIATE, the speaker uses presentation slides that contain the most informative key-phrases. Thus, we extract the text shown in the slides using Google OCR Vision API\footnote{\url{https://cloud.google.com/vision/docs/ocr\#vision\_text\_detection-python}} and fuse the OCR-generated text with ASR-generated text using a novel {\em guided attention mechanism} to attend complementary and non-redundant words of both sources.  

{\bf Guided Attention:} At first, we represent the text in both ASR and OCR-generated transcripts using pre-trained BERT \citep{devlin2018bert}, which provides dynamic embedding for every word. In particular, we use the sequence of 768-dimensional hidden states at the output of the last layer of the BERT model. Let $F\in \mathbb{R}^{n\times 768}$ and $H\in \mathbb{R}^{m\times 768}$ be the BERT representations for ASR and OCR texts respectively, where $n$ and $m$ are the respective token counts. The guided attention mechanism begins with defining an affinity matrix $C\in \mathbb{R}^{n\times m}$, whose element $c_{ij}$ denotes the similarity between the feature vector pairs, $h_i\in \mathbb{R}^{768}$ and $f_i\in \mathbb{R}^{768}$:
\begin{eqnarray}\label{eq_Pm}
 C & = & tanh(H\mathbf{W}^bF^\top)
\end{eqnarray}
where $\mathbf{W}^b\in \mathbb{R}^{768\times 768}$ is a correlation matrix to be learned during training.
 
Subsequently, we compute a normalized weight $\alpha^{h}_{ij}$ to denote the relevance of the $i^{th}$ ASR-generated word to $j^{th}$ OCR-generated word. Therefore, the weighted summation of the ASR transcript, $a_j^h$, can be represented as,  

\begin{equation}\label{eq_Pm}
{a_j^h  =  \sum\limits_{i=1}^n \alpha^{h}_{ij} h_i}
\end{equation}

\begin{equation}
{\text{ where,   } \alpha^{h}_{ij}  =  exp(c_{ij})/\sum\limits_{i=1}^n exp(c_{ij})}
\end{equation}

Since our goal is to emphasize the dissimilar features between the ASR and OCR transcripts, we define the relevance matrix $R(f_i,a_j^h)$ as cosine distance between the attended ASR sentence vector $a_j^h$ and OCR word embedding $f_i$ --
\begin{eqnarray}\label{eq_Pm}
R(f_i,a_j^h) & = & 1 - \dfrac{f_i^\top \cdot a_j^h}{\| f_i\| \| a_j^h\| }
\end{eqnarray}
 
Now, the weighted summation of all word embeddings produces the modified ASR representation $U$ computed as,
\begin{eqnarray}\label{eq_Pm}
U & = & \sum\limits_{j=1}^m R(f_i,a_j^h) \cdot f_i
\end{eqnarray}
where $R(f_i,a_j^h)$ acts as a filter for the ASR encoding $f_i$.

Finally, we concatenate the attended ASR word representations with OCR word embeddings to get the sequential textual features $u_t = \{u_i^t\}_{i = 1}^{l_t}$, which is used as the textual embedding input of FMT.

\subsection{Language Model Pre-training}
The pre-trained Language Model (LM) has recently been shown to have superior performance in abstractive summarization, particularly to enhance sample efficiency \citep{khandelwal2019sample}. This decoder-only network, known as Transformer LM, takes a pre-trained transformer \citep{vaswani2017attention} as its base module and treats summarization as a language modeling task where each generated summary word in every step is appended to its source article. We extend the concept of Transformer LM to a multimodal setting, where we use {\bf Factorized Multimodal Transformer \citep{zadeh2019factorized} based Language Model (\name)} for multimodal sequential learning. After each step of summary generation, we append the generated summary word to its source text transcript, along with a delimiter, and train the transformer on this reformulated data.  \name\ has three crucial advantages over traditional encoder-decoder based summarization networks:

\begin{enumerate}
\item In contrast to encoder-decoder architecture, \name\ uses a single network to encode the source and generate the target, and thus, avoids the loading of same pre-trained weights into separate encoder and decoder. 
\item Compared to the encoder-decoder network, \name\ has fewer number of parameters.
\item Most critically, all the parameters of \name\ can be pre-trained. 
\end{enumerate}

Since there is no available large-scale multimodal corpus, we pre-train \name\ on the text-only 2-billion word corpus\footnote{\url{https://github.com/tensorflow/tensor2tensor}} based on Wikipedia, called WikiLM \citep{khandelwal2019sample}, and fine-tune on the AVIATE and How2  datasets.

\begin{figure*}[h]
    \centering
    \scalebox{0.024}{
    \includegraphics{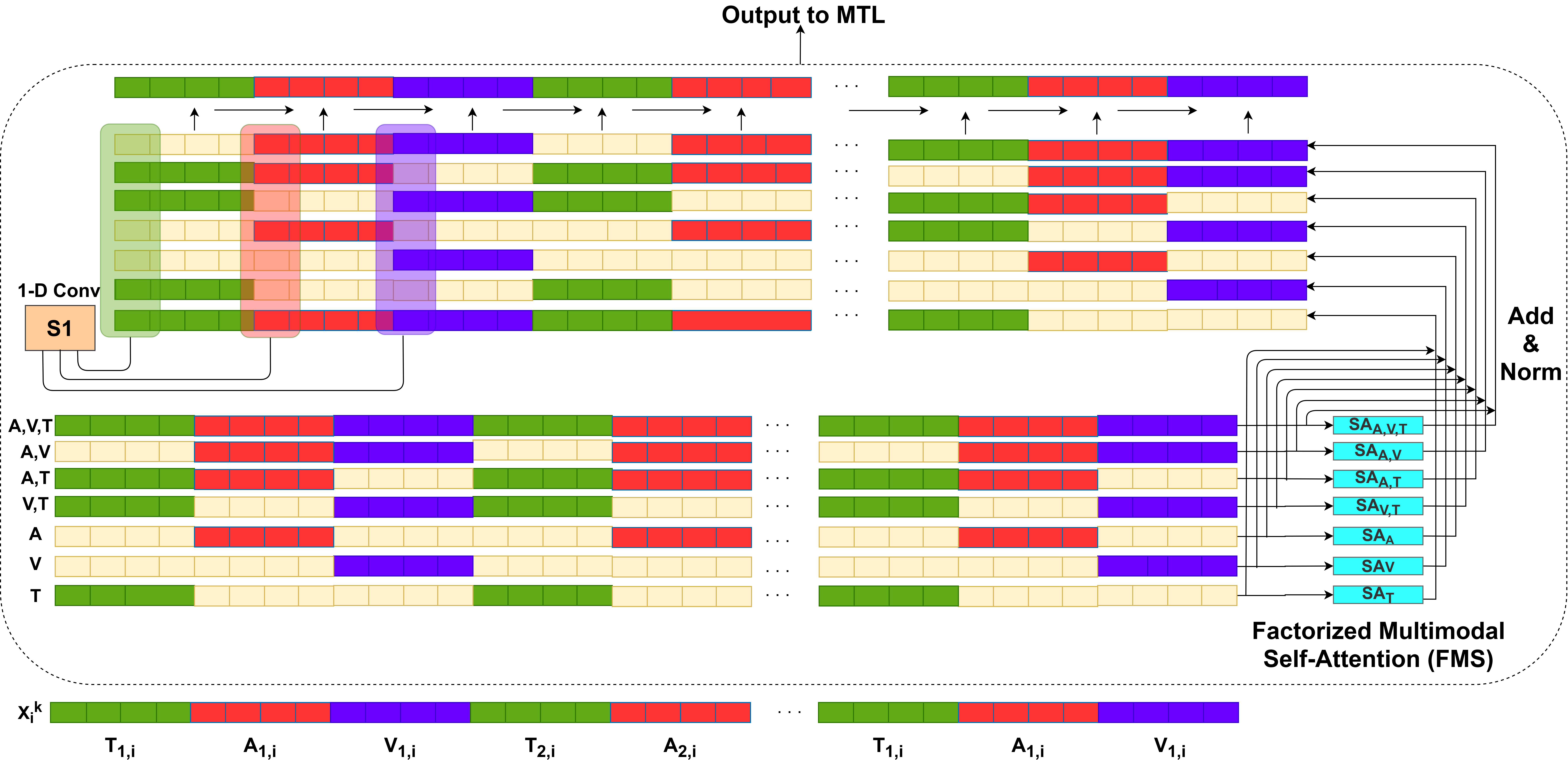}}
    \vspace{-2mm}
    \caption{Overview of a single Factorized Multimodal Self-attention (FMS) in MTL. Each FMS consists of $7$ distinct self-attention \citep{vaswani2017attention} layers, which inherently capture inter-modal and intra-modal dynamics within the asynchronous multimodal input sequence. Blue, red and green colors are used to illustrate the propagation of visual, acoustic and textual features within FMS.}
    \label{fig:fms}
    \vspace{-2mm}
\end{figure*}

\subsection{Factorized Multimodal Transformer LM}

Factorized Multimodal Transformer (FMT) \citep{zadeh2019factorized}, which is the current state-of-the-art model for multimodal emotion recognition and multimodal speaker traits recognition on well-studied IEMOCAP \citep{busso2008iemocap} and POM \citep{park2014computational} datasets, applies seven distinct self-attention mechanisms to simultaneously capture all possible uni-modal, bi-modal, and tri-modal interactions across its multimodal input. We use the FMT architecture as the backbone of our decoder-only Transformer LM. Before feeding into FMT, the unimodal embeddings are resampled using a reference clock so that modalities can follow the same frequency \citep{chen2017multimodal}. Additionally, zero paddings are used to unify the length of all samples of the entire dataset to a desired fixed length $L$. Hence, the $i^{th}$ data point consists of three distinct sequences of embeddings corresponding to three modalities -- visual, acoustic, and language:
\begin{eqnarray}\label{eq_Pm}
D = \{ x_i = [ x_{(t,i)} = \langle u_v^{(t,i)}, u_v^{(t,i)}, u_v^{(t,i)} \rangle ]_{t = 1}^{L}, \: [{tar_i^t}]_{t = 1}^{M} \, \}_{i = 1}^{N}
\end{eqnarray}
where $x_i\in \mathbb{R}^{L\times d_x}$ and ${tar_i\in \mathbb{R}^{M\times d_y}}$ are the inputs and target summaries respectively, $M$ is the length of the summary; $d_x$, $d_y$ denote the input and output dimensionality at each time step respectively; $N$ is the total number of samples within the dataset. Positional embeddings are also added to the input.

The FMT consists of a stack of Multimodal Transformer Layers (MTL), which captures factorized dynamics within multimodal data and aligns the time asynchronous information both within and across modalities using multiple Factorized Multimodal Self-attentions (FMS), each of which has $7$ distinct self-attention layers. Each attention has a unique receptive field with respect to modalities $f \in F = {\tt \{ L,V,A,LV,LA,VA,LVA \}}$. The high dimensional output of FMS is controlled by a summarization network $S_1$ to have a reduced dimension $\mathbb{R}^{L \times d_x}$ which goes through feedforward and normalization layers. If there are a total of $P$ number of FMS units inside MTL, the dimensionality of the normalization layer is $\mathbb{R}^{P \times L \times d_x}$ which is again mapped to $\mathbb{R}^{L \times d_x}$ using a secondary summarization network $S_2$. The output of the last MTL of FMT, thus computed, is fed into a Gated Recurrent Unit (GRU) to have a $d_y$ dimensional predicted summary word embedding, {$Summ_i$}.  An overview of a single Factorized Multimodal Self-attention (FMS) block in MTL is presented in the \ref{fig:fms}.

The summary word predicted by the FMT in the first step, {$Summ_1$}, is appended to the text transcript and fed into the same FMT in the next step to predict the second summary word, {$Summ_2$}. This process is continued until the model generates a stop-word or a predetermined summary length is reached. We only compute loss over the target sequence, as suggested by \cite{khandelwal2019sample}.

%\vspace{-3mm}
\section{Experiments}
To explore the role of multimodality in abstractive text summarization, we conduct multiple experiments evaluating textual and visual modalities separately and jointly on both How2 and AVIATE datasets. Additionally, we investigate the role of OCR-generated text for the academic presentation videos in AVIATE for improving summary generation.

\subsection{Training}
\label{training_params}
We train \name\ using Pytorch framework on NVIDIA Tesla V100 GPU, with 32 GB dedicated memory, with CUDA-10 and cuDNN-7 installed. We pre-train all the parameters of \name\ using WikiLM \citep{khandelwal2019sample}, and fine-tune on the summarization datasets (How2 and AVIATE). Similar to encoder-decoder models, we only compute loss over the target sequence. {In Table \ref{tab:hyper}, we present the details of hyper-parameters used in the baselines and in \name.}

% https://stackoverflow.com/a/49201237 Model parameter counter for pytorch models
% 
\begin{table}[h]
\centering
\resizebox{0.99\textwidth}{!}
{
\begin{tabular}{p{2.5cm} | l | c | c | c | c | c | c }
\multirow{2}{*}{{\bf Modality}} & \multirow{2}{*}{{\bf Models}} & \multicolumn{6}{c}{{\bf Hyperparameters}} \\ \cline{3-8}
& & {\bf Batch-size} & {\bf \#Steps} & {\bf Peak LR}  & {\bf Optimizer}  & {\bf Dropout} &  {\bf \#Parameters} \\ \hline

\multirow{4}{2.5cm} {\centering {\bf Unimodal \\ ({\tt Text} Only)}}
% & {Lead3} &\\
% & {KLSumm} \\
% & {TextRank} \\
% & {LexRank} \\
& {PG} & {16} & {230K} & {0.01} & {Adagrad} & {0.2} & {42m} \\
& {PG-MMR} & {16} & {230K} & {0.01}  & {Adagrad} & {0.2} & {42m} \\
& {Hi-MAP} & {32} & {200K} & {0.01}  & {Adagrad} & {0.1} & {36m} \\
& {CopyTransformer} & {16} & {200K} & {0.05} & {Adam} & {0.2} & {105m} \\

\hline 
\multirow{5}{2.5cm} {\centering {\bf Multimodal \\ ({\tt Text} + {\tt Audio} + {\tt Video})}} & {Multimodal HA} & {64} & {300k} & {0.05} & {Adam} & {0.3} & {8m} \\
& {MulT En-De} & {24} & {300k} & {0.01} & {Adam} & {0.2} & {477m} \\
& {FMT En-De} & {32} & {300k} & {0.01} & {Adam} & {0.2} & {495m}  \\
& {MulT LM} & {32} & {500k} & {0.01} & {Adam} & {0.1} & {242m}  \\
& {\name} & {16} & {500k} & {0.01} & {Adam} & {0.1} & {260m} \\

\hline
\end{tabular}}

\caption{{Hyperparameters of different abstractive baseline models compared to \name.}}

\label{tab:hyper}
\vspace{-2mm}
\end{table}

%%%%%%%%%%%%%%%%%%%%%%%%%%%%%%%%%%  TABLES %%%%%%%%%%%%%%%%%%%%%%%%%%%%%%%%%%
%\input{table_figures/table_all_table}% file containing commands
%\fifthtable
%%%%%%%%%%%%%%%%%%%%%%%%%%%%%%%%%%%%%%%%%%%%%%%%%%%%%%%%%%%%%%%%%%%%%%%%%%%%%

\subsection{Baselines}
We compare the performance of the following extractive and abstractive unimodal and multimodel text summarization models both on How2 and AVIATE datasets.
\subsubsection{Extractive Summarizers (Text Only)}
\begin{itemize}[leftmargin= 0.2in]
\item \textbf{Lead3} is the most common baseline {which simply selects the leading three sentences of the document as its summary.}
\item \textbf{KLSumm} \citep{haghighi-vanderwende-2009-exploring}  is a greedy algorithm that minimizes the Kullback-Lieber (KL) divergence between the original document and the ground-truth summary.
\item \textbf{TextRank} \citep{textrank2004}  runs a modified version of PageRank  on a weighted graph, consisting of nodes as sentences and edges as similarities between sentences.
\item \textbf{LexRank} \citep{Erkan_2004} is a graph-based algorithm that represents sentences as vertices, and edges represent the similarity.
\end{itemize}

\subsubsection{Abstractive Summarizers (Text Only)}
\begin{itemize}[leftmargin= 0.2in]
\item \textbf{Pointer Generator (PG)} \citep{see2017get} network  is one of the most popular sequence to sequence (seq2seq)  summarization architectures. PG allows both generating words from the vocabulary or copying from the source document.

\item \textbf{Pointer Generator-MMR} \citep{lebanoff-song-liu:2018} uses MMR along with PG for better coverage and redundancy mitigation. Here MMR computes a similarity score of sentences with the source text and modifies the attention weights for a better summary generation.

\item \textbf{Hi-MAP} \citep{Multinews2019} is a hierarchical MMR-attention based PG model, which extends the work of PG and MMR. Here, MMR scores are calculated at word level and incorporated in the attention weights for a better summary generation.

\item \textbf{CopyTransformer} (Bottom-up Abstractive Summarization) \citep{gehrmann2018bottom} uses the transformer parameters proposed by \cite{DBLP:journals/corr/VaswaniSPUJGKP17}. It uses a content selection module that over-determine phrases in the source document.
\end{itemize}
 
\subsubsection{Abstractive Summarizers (Video + Audio + Text)}
\begin{itemize}[leftmargin= 0.2in]
\item \textbf{Multimodel Hierarchical Attention} \citep{palaskar2019multimodal} extends the work of \cite{libovicky-helcl-2017-attention}, which was originally proposed for multimodal machine translation. This model fuses visual and textual modalities and captures the context of visual and textual features along with hierarchical attention to generate summaries.
\item \textbf{MulT Encoder-Decoder} is an encoder-decoder based summarization architecture, which uses MulT (Multimodal Transformer for Unaligned Multimodal Language Sequences) model \citep{tsai2019multimodal} as its encoder and decoder unit. 
\item \textbf{FMT Encoder-Decoder} is an encoder-decoder network, similar to MulT encoder-decoder. This baseline uses Factorized Multimodal Transformer {\citep{zadeh2019factorized}} as the encoder and decoder units. 
\item \textbf{MulT LM} is MulT-based  architecture. This is most akin to our proposed \name\ model; only the FMT module of \name\ is replaced by MulT {\citep{tsai2019multimodal}} to have this multimodal summarization baseline.
\end{itemize}

\section{Experimental Results}
\label{section:results}

We present a quantitative analysis of the summaries using the standard metrics for abstractive summarization -- ROUGE-1 (R-1), ROUGE-2 (R-2), and ROUGE-L (R-L) \citep{lin2004rouge, graham2015re} that measure the unigrams, bigrams, and longest common sequence between the ground-truth and the generated summaries, respectively. Additionally, we perform extensive qualitative analysis using human experts to primarily understand the fluency and informativeness of the summaries. We also analyze the word distributions in the transcriptions and summaries.

%%%%%%%%%%%%%%%%%%%%%%%%%%%%%%%%%%  TABLES  %%%%%%%%%%%%%%%%%%%%%%%%%%%%%%%%%%%%%%%%%%%%%%
% \input{table_figures/table_all_table}% file containing cammands

% \firsttable
\begin{table}[!t]
\centering
\small
\setlength{\extrarowheight}{0.8pt}

% \resizebox{0.9999\columnwidth}{!}{
\resizebox{0.75\textwidth}{!}{
\begin{tabular}{p{2cm}|p{2.8cm}|p{2.9cm}|P{0.9cm}|P{0.9cm}|P{0.9cm}}

& \centering \multirow{2}{*}{\bf Model} & \centering \multirow{2}{*}{\bf Modality} & \multicolumn{3}{c}{\bf AVIATE Dataset} \\\cline{4-6}

& & & \bf R-1 & \bf  R-2 & \bf R-L\\
\hline

\multirow{6}{2cm}{Extractive {\tt (Text} only)} & \multirow{2}{1.8cm}{KLSumm} & {\tt ASR} & 22.19 & 2.05 & 15.59\\
& & {\tt ASR}+{\tt OCR} & 24.27 & 2.31 & 16.92 \\ \cline{2-6}

& \multirow{2}{1.8cm}{TextRank} & {\tt ASR} & 22.15 & 2.71 & 16.42\\
& & {\tt ASR}+{\tt OCR} & 24.55 & 2.72 & 22.1  \\ \cline{2-6}

& \multirow{2}{1.8cm}{LexRank} & {\tt ASR} & 22.63 & 2.49 & 15.68\\
& & {\tt ASR}+{\tt OCR} & 24.55 & 2.72 & 22.1 \\
\hline

\multirow{8}{2cm}{Abstractive {\tt (Text} only)} & \multirow{2}{1.8cm}{PG} & {\tt ASR} & 26.27 & 2.01 & 22.96\\
& & {\tt ASR}+{\tt OCR} & 27.77 & 2.05 & 23.81  \\ \cline{2-6}

& \multirow{2}{1.8cm}{PG-MMR} & {\tt ASR} & 27.34 & 2.72 & 22.63\\
& & {\tt ASR}+{\tt OCR} & 27.82 & 3.97 & 23.93 \\ \cline{2-6}

& \multirow{2}{1.8cm}{Hi-MAP} & {\tt ASR} & 27.62 & 3.16 & 22.1\\
& & {\tt ASR}+{\tt OCR} & 28.13 & 3.87 & 22.5\\ \cline{2-6}

& \multirow{2}{1.8cm}{CopyTransformer} & {\tt ASR} & 29.93 & 3.73 & 25.13\\
& & {\tt ASR}+{\tt OCR} & 30.27 & 3.94 & 27.06\\

\hline
\hline 

\multirow{10}{2cm}{Multimodal  {\tt (Text} + {\tt Audio} + {\tt Video)}} & \multirow{2}{2.8cm}{Multimodal HA} & {\tt ASR}+{\tt A}+{\tt V} & 27.51 & 4.83 & 25.32 \\
& & {\tt (ASR}+{\tt OCR)}+{\tt A}+{\tt V} & 28.14 & 4.91 & 26.12 \\ \cline{2-6}

& \multirow{2}{2.8cm}{MulT Encoder-Decoder} & {\tt ASR}+{\tt A}+{\tt V} & 29.65 & 4.12 & 26.47 \\
& & {\tt (ASR}+{\tt OCR)}+{\tt A}+{\tt V} & 30.89 & 4.34 & 27.2 \\ \cline{2-6}

& \multirow{2}{2.8cm}{FMT Encoder-Decoder} & {\tt ASR}+{\tt A}+{\tt V} & 31.8 & 4.49 & 26.1 \\
& & {\tt (ASR}+{\tt OCR)}+{\tt A}+{\tt V} & 32.85 & 4.6 & 27.65 \\ \cline{2-6}

& \multirow{2}{2.8cm}{{MulT LM}} & {{\tt ASR}+{\tt A}+{\tt V}} & {31.71} & {4.07} & {27.58} \\
& & {{\tt (ASR}+{\tt OCR)}+{\tt A}+{\tt V}} & {33.47} & {4.12} & {28.73}
 \\ \cline{2-6}

& \multirow{2}{2.8cm}{\name} & {\tt ASR}+{\tt A}+{\tt V} & 33.26 & 6.38 & 28.52 \\
& & {\tt (ASR}+{\tt OCR)}+{\tt A}+{\tt V} & {\bf 37.13} & {\bf 11.04} & {\bf 31.47} \\

\hline

\end{tabular}}
% }
% \vspace{0.1cm}
\caption{Ablation results after incorporating  OCR generated text into the ASR generated text transcript using guided-attention for different extractive and abstractive unimodal and multimodal text summarization systems on AVIATE.}
\label{tab:ocr}
%\vspace{-10mm}
\end{table}
%%%%%%%%%%%%%%%%%%%%%%%%%%%%%%%%%%%%%%%%%%%%%%%%%%%%%%%%%%%%%%%%%%%%%%%%%%%%%%%%%%%%%%%%%%%

\subsection{Quantitative Analysis}
At first, we evaluate the performance of commonly used extractive and abstractive text summarization models both on the How2 and AVIATE datasets. Note that the average length of text transcripts in How2 is much less than that of AVIATE. Following our intuition, PG-based text summarization networks perform relatively well on How2 as shown in Table \ref{tab:compariosn}; but their performance drastically drops on AVIATE. This result can be attributed to the fact that attention-based encoder-decoder networks often fail to capture long-term dependencies when the source text is long and noisy. Hence, we decide to use transformer-based pre-trained BERT \citep{devlin2018bert} as the text-embedding layer in our model.

In addition to text-only models, we train two video-only models -- the first one uses a single convolutional and pooling layer for feature extraction from the entire video, while the second one applies a single layer RNN over these vectors in time. We observe in Table \ref{tab:compariosn} that even using only action features in the videos leads to almost competitive R-1, R-2, and R-L scores compared to text-only models, in some cases often better than extractive text-only systems. This result demonstrates the importance of both modalities for summarization.

%%%%%%%%%%%%%%%%%%%%%%%%%%%%%%%%%%  TABLES  %%%%%%%%%%%%%%%%%%%%%%%%%%%%%%%%%%%%%%%%%%%%%%
% \input{table_figures/table_all_table}% file containing cammands
% \secondtable

\begin{table*}[hbtp]
\centering % centering table

\resizebox{0.9\textwidth}{!}
{
\begin{tabular}{p{3.9cm}|p{3.6cm}|P{1cm}|P{1cm}|P{1cm}|P{1cm}|P{1cm}|P{1cm}}
% \hline
\multirow{2}{*}{\bf Modality} & \multirow{2}{*}{\bf Model} & \multicolumn{3}{c|}{\bf How2 } & \multicolumn{3}{c}{\bf AVIATE } \\\cline{3-8}

& & \bf R-1 & \bf  R-2 & \bf R-L & \bf  R-1 & \bf R-2 & \bf  R-L\\ 

\hline
\hline

\multirow{4}{3cm}{Extractive Systems {\tt (Text} only)} & Lead3 & 43.97 & 10.31 & 35.76 & 19.23 & 1.82 & 14.69 \\
& KLSumm & 28.64 &	\centering 12.3	& 16.2 & 24.27 & 2.31 & 16.92 \\
& TextRank & 27.48 & 12.41 & 16.55 & 22.15 & 2.71 & 16.42 \\
& LexRank & 27.95 & 12.48 & 16.98 & 22.63 & 2.49 & 15.68 \\

\hline

\multirow{5}{3cm}{Abstractive Systems {\tt (Text} only)} & Seq2Seq  & 55.32 & 23.06 & 53.9 & 26.62 & 2.88 & 23.32 \\
& PG & 51.67 & 22.65 & 50.21 & 27.77 & 2.05 & 23.81 \\
& PG-MMR & 52.9 & 23.21 & 50.24 & 27.34 & 2.72 & 22.63 \\
& Hi-MAP & 49.2 & 21.36 & 47.36 & 28.13 & 3.87 & 22.5\\
& CopyTransformer & 53.7 & 23.87 & 48.04 & 30.27 & 3.94 & 27.06 \\
\hline

\multirow{2}{3.2cm}{{\tt Video} only} & Action Ft. only & 45.21 & 12.74 & 38.5 & 22.87 & 2.78 & 17.13 \\
& Action Ft. + RNN & 48.23 & 19.1 & 46.3 & 23.54 & 2.92 & 18.21 \\

\hline
\hline

\multirow{5}{3.5cm}{Multimodel Systems {\tt (Video}+{\tt Audio}+{\tt Text)}} & Multimodal HA & 55.87 & 26.32 & 54.9 & 28.14 & 4.91 & 26.12 \\
& MulT En-De & 55.89 & 26.79 & 55.1 & 30.89 & 4.34 & 27.2 \\
& FMT En-De & 55.98 & 26.83 & 55.4 & 32.85 & 4.6 & 27.65 \\
& MulT LM & 56.13 & 26.89 & 55.41 & 33.47 & 4.12 & 28.73 \\
& \textbf{\name} & \bf 56.89 & \bf 26.93 & \bf 56.80 & \bf 37.13 & \bf 11.04 & \bf 31.47 \\

\hline

\end{tabular}}
%\vspace{0.1cm}
\caption{\name\ achieves highest performance in ROUGE-1, ROUGE-2 and ROUGE-L over text based extractive system (Lead3, KLSumm, TextRank and LexRank) and abstractive systems (Seq2Seq, PG, PG-MMR, Hi-MAP and CopyTransformer) and multimodel baselines (Multimodel HA, MulT based encoder decoder, FMT based encoder decoder, MulT based language model and \name) in How2 and AVIATE datasets.}
\label{tab:compariosn}
%\vspace{-6mm}
\end{table*}
%%%%%%%%%%%%%%%%%%%%%%%%%%%%%%%%%%%%%%%%%%%%%%%%%%%%%%%%%%%%%%%%%%%%%%%%%%%%%%%%%%%%%%%%%%%

\subsubsection{\textit{\textbf{Incorporation of OCR}}} 
Since the AVIATE dataset is composed of conference presentation videos, we observe that in almost $94.8\%$ videos in the entire dataset, the speaker shows slides during the presentation. The text in these slides is succinct and contains the most important key-phrases which are crucial for summary generation. Table \ref{tab:ocr} shows the performance improvement for every summarization model when the OCR is fused with ASR transcript using our guided attention mechanism. We also consider direct concatenation of OCR transcript with the ASR-transcript; however, it resulted in lower performance as compared to guided attention fusion. The guided attention ensures the filtering of redundant and repetitive words in OCR and ASR transcripts. For every summarization model, we only use the first $500$ tokens of the OCR transcript. We did not consider incorporating OCR for How2 as this dataset only contains instructional videos, and there is no text shown in the frames of instructional videos. 

Table \ref{tab:ocr} shows that unimodal extractive text summarization models, namely KLSumm, TextRank and LexRank, yield an improvement of $[2.1-2.3]$ R-1 points, $[0.1-0.3]$ R-2 points and $[0.6-6.5]$ R-L points after incorporating OCR-generated text. Similarly for abstractive summarization, the very popular PG-MMR network produces $0.5, 1.25$, and $2.3$ points performance improvement in terms of R-1, R-2, and R-L scores, respectively. The other abstractive summarization networks, namely PG, Hi-MAP, CopyTransformer, also support our hypothesis and show an improvement of $[0.5-4]$ points in terms of all three evaluation metrics. 

Influenced by the performance of unimodal summarization models, we incorporate the OCR transcripts into all of our multimodal baselines. Supporting our intuition, the multimodal systems obtain significant performance enhancement with OCR transcripts as shown in Table \ref{tab:ocr}. The multimodal hierarchical attention model, MulT, and FMT-based encoder-decoder models show $[0.8-1.5]$ points improvement in the R-L score. Our proposed \name\ model yields the highest performance boost with OCR among all the multimodal systems, showing $3.87$, $4.66$, and $2.95$ point enhancement in R-1, R-2, and R-L scores respectively. The performance boost can be easily attributed to the keywords in the OCR-generated transcript, which guides the text-embeddings to attend the most important portions in a very long ASR transcript. Hence, in the rest of our discussion, we always report results with {\tt (ASR} + {\tt OCR)} transcript, fused with guided attention, as the textual modality.

\subsubsection{\textit{\textbf{Complementarity of Multiple Modalities}}} 
Table \ref{tab:compariosn} shows the ROUGE scores for different unimodal and multimodal text summarization systems on the How2 and AVIATE datasets. Among the unimodal variants, the abstractive text summarization systems generally perform much better than the extractive systems, especially on AVIATE. Note that despite being a very strong extractive baseline, Lead3 does not perform well on AVIATE, as the text transcripts of academic presentation videos do not tend to be structured with the most important information at the beginning. The two video-only models, simple conv-pool action features and action features with RNN perform very close to the abstractive text-only baselines, which clearly indicates the necessity of visual modality in addition to the textual modality.\footnote{We do not evaluate the performance of acoustic modality separately, as the MFCC audio features are typically incorporated to capture the pitch, intonation, and other tonal-specific details of the speaker, which do not contribute individually to the summarization task.} As presented in Table \ref{tab:compariosn}, the MulT, and FMT multimodal baselines and the proposed \name\ model beat most of the unimodal systems by a large margin, on both the datasets. This result is expected because of the inherent ability of MulT and FMT to capture the intra-model and inter-modal dynamics within asynchronous multimodal sequences and incorporate diverse information in a single network. Overall, the combination of visual, acoustic, and textual signals significantly improves over the unimodal variants, with an improvement of $1.57$, $3.04$, and $3$ R-1, R-2, and R-L points on How2 and $6.86$, $7.1$ and $4.41$ on AVIATE.

\begin{figure*}[t!]
\hspace*{0cm}
  \includegraphics[width=0.98\textwidth]{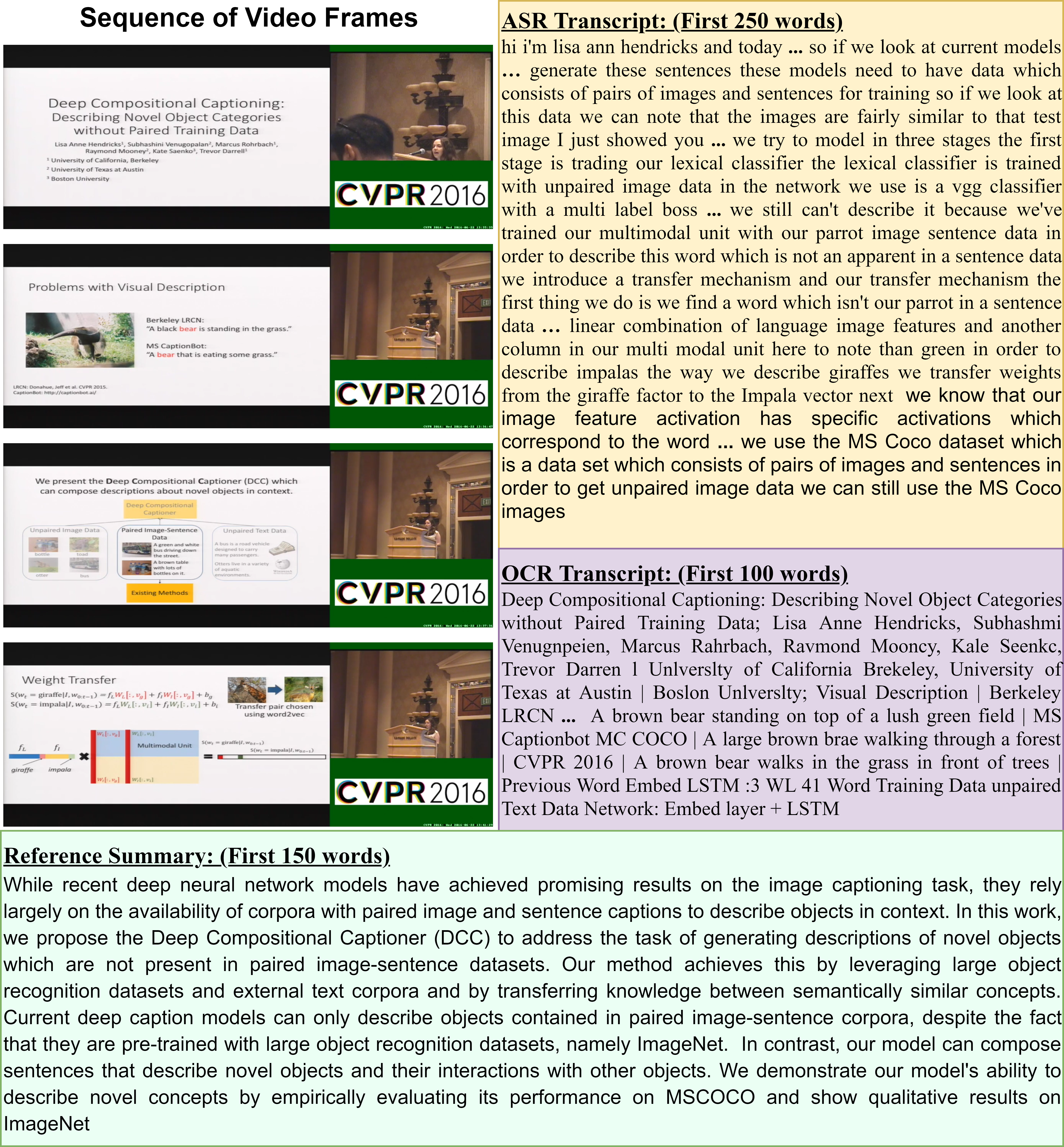}
  \caption{{Example of AVIATE dataset with three different modalities. To obtain the text transcripts from the acoustic modality, we apply Deep Speech \citep{DBLP:journals/corr/HannunCCCDEPSSCN14}, a pre-trained end-to-end automatic speech recognition (ASR) system. We extract the text shown in the slides in the presentation videos using Google OCR Vision API. We use the abstracts of corresponding research papers as the ground-truth summaries.}}
  \label{fig:example_aviate}
  \vspace{-5mm}
\end{figure*}

We manually investigate some video samples of AVIATE where the multimodal system generates a better summary than the unimodal system. In most of these samples, the textual transcript is very noisy and contains many irrelevant words that are not much required for the summary generation. {Figure \ref{fig:example_aviate} shows an example training instance of the AVIATE dataset with three different modalities.} A closer look into the ASR and OCR transcripts reveals the presence of irrelevant and noisy words. For example, the very first sentence of the ASR transcript ”hi i’m lisa ann hendricks and today” does not contribute to the summary generation. As a result, these samples require additional cues for performance improvement, which are availed from the multimodal signals. The variation of outputs from various unimodal and multimodal summarization networks for a single video sample is shown in Table \ref{tab:output_examples}.

\subsubsection{\textit{\textbf{Comparative Study on How2}}} Table \ref{tab:compariosn} shows that the performance of unimodal and multimodal summarization systems on How2 as compared to AVIATE. In contrast to prior work on news-domain summarization \citep{nallapati2016abstractive}, the seq2seq model performs the best among all unimodal systems on  How2, achieving $55.32$, $23.06$, and $53.9$ R-1, R-2, and R-L scores, respectively. As indicated by \cite{palaskar2019multimodal}, the PG model performs lower than seq2seq on How2 due to the lack of overlaps between input and output, which is the important feature of PG networks. Among the multimodal systems, our proposed \name\ model yields the best results; however, the other multimodal baselines reach almost competitive ROUGE scores compared to \name\ on this dataset. Noticeably, despite having a simple structure, the multimodal hierarchical attention model performs very well on How2. On this dataset, \name\ achieves $56.89$, $26.93$, and $56.80$ R-1, R-2, and R-L scores, respectively, which are $[0.1-2]$ points higher than the scores achieved by other multimodal baselines.

\subsubsection{\textit{\textbf{Comparative Study on AVIATE}}} AVIATE contains longer videos than How2, resulting in longer transcripts and ground-truth summaries. As shown in Table \ref{tab:compariosn}, the best performing unimodal summarization model on AVIATE is CopyTransformer. As the ASR and OCR generated summaries are very long, the extractive systems do not perform well on this dataset. While PG and seq2seq yield $23.32$ and $23.81$ R-L scores respectively, CopyTransformer produces a $27.06$ R-L score, outperforming all other unimodal systems. The superior performance of CopyTransformer over PG and seq2seq can be attributed to the self-attention mechanism of transformers which helps to capture long-term dependencies. The incorporation of visual and acoustic modalities significantly improves the ROUGE scores on this dataset. \name\ beats all the transformer-based encoder-decoder networks and language models.  \name\ produces $37.13$, $11.04$ and $31.47$ R-1, R-2 and R-L scores respectively, where the second-ranked model on AVIATE, {\tt MulT-LM} obtains $33.47$ R-1, $4.12$ R-2, $28.73$ R-L scores, which are almost $[2.7-7]$ points lower than that of \name. Since AVIATE contains $6,680$ training samples, which may not be enough for today's deep neural models, the factorization mechanism on FMT, which allows an increasing number of self-attention to better model the multimodal phenomena, results in its superior performance, without encountering difficulties even on the relatively low-resource setup of AVIATE. Pre-training of all the parameters of \name\ also has an immense impact, which helps in beating all other baselines by a significant margin.

\begin{table}[b]
\centering
\small
\setlength{\extrarowheight}{0.8pt}

% \resizebox{0.9999\columnwidth}{!}{
\resizebox{0.95\textwidth}{!}{
\begin{tabular}{p{2.7cm}|p{0.9cm}|P{0.9cm}|P{0.9cm}|P{0.9cm}|P{0.9cm}|P{0.9cm}|P{0.9cm}|P{0.9cm}|P{0.9cm}|P{0.9cm}|P{0.9cm}|P{0.9cm}}

\multirow{3}{*}{{\bf Model}} & \multicolumn{12}{c}{{\bf AVIATE Dataset}} \\ \cline{2-13}

& \multicolumn{3}{c|}{\bf {Short Videos}} & \multicolumn{3}{c|}{\bf {Medium Videos}} & \multicolumn{3}{c|}{\bf {Long Videos}} & \multicolumn{3}{c}{\bf {Whole Dataset}} \\ \cline{2-13}

& \bf {R-1} & \bf  {R-2} & \bf {R-L} & \bf {R-1} & \bf {R-2} & \bf {R-L} & \bf {R-1} & \bf {R-2} & \bf {R-L} & \bf {R-1} & \bf {R-2} & \bf {R-L}\\ \cline{2-13}
\hline

{Multimodal HA} & {30.11} & {5.12} & {27.03} & {26.54} & {4.97} & {25.31} & {24.18} & {4.65} & {23.19} & {28.14} & {4.91} & {26.12}
\\
{MulT En-De} & {31.44} & {5.32} & {27.38} & {27.73} & {4.93} & {26.89} & {25.62} & {4.67} & {24.12} & {30.89} & {4.34} & {27.2}\\ 
{FMT En-De} & {33.62} & {5.81} & {31.06} & {32.1} & {5.91} & {28.31} & {31.48} & {4.96} & {27.02} & {32.85} & {4.6} & {27.65}\\
{MulT LM} & {33.13} & {5.65} & {30.58} & {34.09} & {5.95} & {29.27} & {33.37} & {5.31} & {28.69} & {33.47} & {4.12} & {28.73}
\\ 
{\name} & {36.13} & {11.62} & {31.44} & {35.59} & {11.64} & {31.05} & {34.31} & {10.39} & {30.8} & {37.13} & {11.04} & {31.47}
\\

\hline

\end{tabular}}
% }
% \vspace{0.1cm}
\caption{{Performance of multimodal baseline models and \name\ on short ($<10$ min), medium ($>10$ min \& $<30$ min) and long ($>30$ min) videos of AVIATE. As the video length and the corresponding reference summary length increase, the performance of all baseline models decreases heavily. However, \name\ performs well across all video lengths.}}
\label{tab:short_long_medium}
%\vspace{-10mm}
\end{table}

\begin{table*}[h]
\centering % centering table

\resizebox{0.5\textwidth}{!}
{
\begin{tabular}{ l | c | c | c | c}
% \hline
\multirow{2}{*}{\bf {Model}} & \multirow{2}{*}{\bf {Modality}} & \multicolumn{3}{c}{\bf {AVIATE}} \\ \cline{3-5}

& & \bf {R-1} & \bf {R-2} & \bf {R-L} \\

\hline
\hline

\multirow{4}{*}{{\name}} & {\tt {T}}  & {\bf 27.14} & {\bf 2.93} & {\bf 19.67} \\
& {\tt {A}}  & {21.64} & {1.97} & {16.83} \\
& {\tt {V}} & {22.31} & {2.29} & {17.04} \\ \cline{2-5}

& {\tt {T+A+V}} & {\bf 37.13} & {\bf 11.04} & {\bf 31.47}\\ \cline{1-5}

\multicolumn{2}{c|}{\bf {$\Delta_{multi-unimodal}$}} & \textcolor{green!60!black}{$\uparrow$ 9.99} & \textcolor{green!60!black}{$\uparrow$ 8.11} & \textcolor{green!60!black}{$\uparrow$ 11.80}  \\

\hline

\end{tabular}}
%\vspace{0.1cm}
\caption{{Significance of multimodal cues in \name. The combination of visual, textual, and acoustic signals significantly improves over the unimodal variants, with a relative improvement of R-$1$, R-$2$ and R-L scores of $9.99\%$, $8.11\%$ and $11.80\%$ respectively over the best unimodal variant. }}
\label{tab:floral_multi_unimodal}
%\vspace{-5mm}
\end{table*}

Table \ref{tab:compariosn} also shows that all the unimodal and multimodal summarization models obtain almost $[18-25]$ points higher R-1 and R-L scores and $[3-6]$ points higher R-2 score on How2 over AVIATE. For example, \name\ yields $56.89$, $26.93$, and $56.80$ R-1, R-2, and R-L scores on How2 and $37.13$, $11.04$ and $31.47$ R-1, R-2 and R-L scores on AVIATE. {We can observe from Table \ref{tab:compariosn} that all the baseline models as well as \name\ yield higher R-1, R-2 and R-L scores on How2 than AVIATE.} The overall better performance of every system on How2 than AVIATE can be attributed to two factors -- firstly, the text transcripts of How2 are manually annotated. In contrast, we use ASR and OCR outputs as the transcripts for AVIATE. The large margin of ASR and OCR errors in some of the train and test samples significantly affect the model performance. Secondly, {since the video length, transcript length, and reference summary length are much longer in AVIATE than How2, the summarization task becomes more challenging in AVIATE. Furthermore, since AVIATE comprises many scientific presentation videos, the audio transcript contains complex academic words, leading to a larger dictionary for the language generation task. Overall, the results in Table \ref{tab:compariosn} conclude that the AVIATE dataset is more exacting than How2, indicating room for further research with fine-grained and sophisticated multimodal models for long videos.} 

{In our next experiment, we demonstrate how the summarization task becomes more challenging with longer videos. We divide the AVIATE dataset into three portions - short videos (duration less than $10$ minutes), medium videos (duration between $10$ minutes and $30$ minutes) and long videos (duration more than $30$ minutes). We split each portion in $4:1$ ratio and train and test all the multimodal systems on each segment. Table \ref{tab:short_long_medium} shows the performance reduction of each model with the increase in video length. In general, we observe that all four multimodal baselines yield R-L score in the range of $[27.03 - 34.09]$ on the short videos. However, the score reduces to $[23.19 - 28.69]$ for the long videos. The performance of \name\ also decreases from short to medium and long videos; however, the span of reduction of R-L score is only $[0.39 - 0.64]$, which is relatively less than all other baselines. We also notice that the LM-based systems generally capture long-term dependencies better than traditional encoder-decoder based systems.}

{The complementarity of multiple modalities in the performance of \name\ is shown in Table \ref{tab:floral_multi_unimodal}. To understand the importance of visual modality, we feed zero input in other two modality channels of \name\ and continue the process for all three modalities. We observe that the textual modality provides the best performance among unimodal variants. The addition of visual and acoustic features improves significantly over the unimodal baselines and achieves the best performance - with an increase in R-$1$, R-$2$ and R-L score of $9.99$, $8.11$ and $11.80$ respectively over the best unimodal variant.}

% \input{table_figures/table_all_table}% file containing cammands
% \fourthtable

\begin{table}[hbtp]
\centering

\setlength{\extrarowheight}{0.8pt}

\resizebox{0.7\textwidth}{!}{

\begin{tabular}{c|c|c|c|c|c|c}

& \multicolumn{3}{c|}{\bf How2} & \multicolumn{3}{c}{\bf AVIATE} \\\cline{2-7}
& \bf R-1 & \bf  R-2 & \bf R-L & \bf R-1 & \bf  R-2 & \bf R-L \\
\hline

\bf How2 & 56.89 & 26.93 & 56.80 & 21.35 & 4.88 & 23.11 \\
\hline 
\bf How2 with ASR & 54.19 & 24.07 & 52.11 & 22.77 & 5.98 & 24.07
 \\
\hline
\bf AVIATE & 52.68 & 21.33 & 49.9 & 37.13 & 11.04 & 31.47 \\

\hline

\end{tabular}}
%\vspace{0.1cm}
\caption{Transferability of the proposed \name\  model on the two available multimodal abstractive text summarization datasets - How2 and AVIATE. The network is trained on the dataset in each row, and is tested on the dataset shown in each
column. The second row indicates the performance of \name\ on the How2 videos whose transcripts are generated from ASR.}
\label{tab:transferability}
%\vspace{-8mm}
\end{table}

\begin{comment}

\begin{table}[hbtp]
\centering

\setlength{\extrarowheight}{0.8pt}

\resizebox{0.7\textwidth}{!}{

\begin{tabular}{c|c|c|c|c|c|c}

& \multicolumn{3}{c|}{\bf How2} & \multicolumn{3}{c}{\bf AVIATE} \\\cline{2-7}
& \bf R-1 & \bf  R-2 & \bf R-L & \bf R-1 & \bf  R-2 & \bf R-L \\
\hline

\bf How2 & 54.19 & 24.07 & 52.11 & 19.67 & 3.54 & 21.87
 \\
\hline
\bf AVIATE & 52.68 & 21.33 & 49.9 & 37.13 & 11.04 & 31.47 \\

\hline

\end{tabular}}
%\vspace{0.1cm}
\caption{Transferability of the proposed \name\  model when instead of the annotated transcript, Automated method for Transcription generation \citep{DBLP:journals/corr/HannunCCCDEPSSCN14} is used on the two available multimodal abstractive text summarization datasets  - How2 and AVIATE. The network is trained on the dataset in each row and is tested on the dataset shown in each column.}
\label{tab:transferability}
%\vspace{-8mm}
\end{table}

\end{comment}

\begin{table}[hbtp]
\centering

\setlength{\extrarowheight}{0.8pt}

\resizebox{0.90\textwidth}{!}{
\begin{tabular}{c|c|p{0.9cm}|p{0.9cm}|p{0.9cm}|p{0.9cm}|p{0.9cm}|p{0.9cm}|p{0.9cm}|p{0.9cm}|p{0.9cm}}

& & \multicolumn{9}{c}{\bf {AVIATE}} \\\cline{3-11}
& & \multicolumn{3}{c|}{\bf {Short Videos}} & 
 \multicolumn{3}{c|}{\bf {Medium Videos}} & \multicolumn{3}{c}{\bf {Long Videos}} \\ \cline{3-11}
& & \bf {\centering R-1} & \bf  {R-2} & \bf {R-L} & \bf {R-1} & \bf {R-2} & \bf {R-L} & \bf {R-1} & \bf {R-2} & \bf {R-L} \\

\hline

\bf \multirow{3}{*}{{AVIATE}} & \bf {Short Videos} & {31.05} & {8.49} & {28.25} & {27.22} & {7.36} & {23.74} & {24.71} & {6.12} & {21.53} \\ \cline{2-11}

& \bf {Medium Videos} & {31.43} & {8.74} & {28.34} & {30.21} & {8.48} & {26.90} & {28.66} & {8.29} & {26.54}
 \\ \cline{2-11}

& \bf {Long Videos} & {32.84} & {9.43} & {28.56} & {31.08} & {9.64} & {26.98} & {30.23} & {8.89} & {27.25} \\ \cline{2-11}

\hline

\end{tabular}}
%\vspace{0.1cm}
\caption{Transferability of the proposed \name\ model on videos of different length in the AVIATE dataset. The network is trained on the videos in each row, and tested on the videos shown in each column. }
\label{tab:transferability_short_medium_long}
%\vspace{-8mm}
\end{table}

\subsubsection{\textit{\textbf{Transferability of \name}}}
Table \ref{tab:transferability} shows the transferability property of \name\ between How2 and AVIATE. When trained and tested on the same dataset, \name\ produces the best ROUGE scores, which is expected. However, when trained on AVIATE and tested on How2, \name\ yields an R-L score of $49.90$, which is just $6.9$ decrease in R-L score {($11.83\%$ reduction in performance)} than the one when trained and tested on How2. The vice-versa is not true, i.e., when trained on How2 and tested on AVIATE, the performance drop is drastic {($26.56\%$ reduction in performance)}. As the videos in How2 have human-annotated transcripts and those in AVIATE have ASR-generated transcripts, for fair comparison of transferability, we extract the ASR transcripts of the How2 videos and train \name. The results of this experiment are shown in the second row of Table \ref{tab:transferability}. We observe that the ASR transcript reduces the test performance on How2, which is expected due to the noise in the ASR output. The transferability score on AVIATE improves a bit, but the performance drop is still heavy {($25.51\%$ reduction in performance)}. From all these experiments, we can conclude that since the videos of How2 are very short, the learned weights do not perform well for longer videos. However, AVIATE consists of diverse-length videos, and thus, the trained model on AVIATE yields good results on How2 as well.

{Table \ref{tab:transferability_short_medium_long} shows the transferability of \name\ across short, medium and long videos of AVIATE. When trained on the long videos, \name\ performs the best across all three portions. However, when trained on short videos, the model can not learn long-term dependency for lengthier videos. The same property supports the results on Table \ref{tab:transferability}. Since the How2 dataset contains only short videos, the model does not perform well when trained on How2 and tested on AVIATE. The longer videos in the training set helps the model to generalize well across videos of various lengths. }

%%%%%%%%%%%%%%%%%%%%%%%%%%%%%%%%%%%%%%%%%%%%%%%%%%%%%%%%%%%%%%%%%%%%%%%%%%%%%%%%%%%%%%%%%%%%
% \input{table_figures/table_all_table}% file containing cammands
% \thirdtable

\begin{table*}[hbtp]

\centering % centering table

\resizebox{1\textwidth}{!}
{
\begin{tabular}{p{4.2cm}|p{2.2cm}|P{0.9cm}|P{0.9cm}|P{0.9cm}|P{0.9cm}|P{0.9cm}|P{0.9cm}| P{0.9cm}|P{0.9cm}}
% \hline
\multirow{2}{*}{\bf Modality} & \multirow{2}{*}{\bf Model} & \multicolumn{4}{c|}{\bf How2 } & \multicolumn{4}{c}{\bf AVIATE } \\\cline{3-10}

& & \bf INF & \bf  REL & \bf COH & \bf FLU & \bf INF & \bf  REL & \bf COH & \bf FLU \\ 

\hline
\hline

\multirow{2}{3.5cm}{Extractive Systems {\tt \footnotesize{(Text only)}}} & KLSumm & 2.82 & 2.54 & 2.98 & 3.14 & 1.91 & 1.56 & 2.14 & 2.13  \\
& TextRank & 2.92 & 2.73 & 2.82 & 3.12 & 2.1 & 1.83 & 2.13 & 2.12   \\

\hline

\multirow{2}{3.5cm}{Abstractive Systems \footnotesize{(Text only)}} &  PG & 3.45 & 3.17 & 3.12 & 3.32 & 3.49 & 3.38 & 3.41 & \bf 3.62 \\
& CopyTrans. & 3.46 & 3.18 & 3.18 & 3.36 & 3.54 & 3.39 & 3.48 & 3.56 \\
\hline

\multirow{1}{3.5cm}{{Abstractive Systems \footnotesize{(Video only)}}}  & Action Ft.+RNN & \multirow{2}{*}{3.54} & \multirow{2}{*}{3.20} & \multirow{2}{*}{3.21} & \multirow{2}{*}{3.40} & \multirow{2}{*}{3.52} & \multirow{2}{*}{3.27} & \multirow{2}{*}{3.31} & \multirow{2}{*}{3.41} \\

\hline
\hline

\multirow{3}{3.5cm}{Multimodel Systems \footnotesize{(Video+Audio+Text)}} & FMT En-De & 3.61 & \bf 3.39 & 3.37 & 3.67 & 3.64 & 3.32 & 3.37 & 3.49  \\
& MulT LM & 3.57 &  3.38 & 3.34 & 3.68 & 3.67 & 3.39 & 3.38 & 3.45 \\
& \textbf{\name} & \bf 3.62 & 3.38 & \bf 3.41 & \bf 3.71  & \bf 3.89 & \bf 3.41 & \bf 3.41 & 3.41 \\

\hline

\end{tabular}}
%\vspace{0.1cm}
\caption{Scores for human evaluated metrics - Informativeness (INF), Relevance (REL), Coherence (COH), Fluency (FLU) over text based extractive systems (KLSumm and TextRank), abstractive systems (PG and CopyTransformer), video based abstractive systems (Action features with RNN) and multimodel systems (FMT Encoder Decoder, MulT Language Model and \name\ ) on How2 and AVIATE datasets.}
\label{tab:human}
%\vspace{-6mm}
\end{table*}
%%%%%%%%%%%%%%%%%%%%%%%%%%%%%%%%%%%%%%%%%%%%%%%%%%%%%%%%%%%%%%%%%%%%%%%%%%%%%%%%%%%%%%%%%%%%

\begin{figure*}[!t]
  \centering
  \scalebox{0.88}{
  \subfigure[Density curve on How2 dataset.]{\includegraphics[scale=0.40]{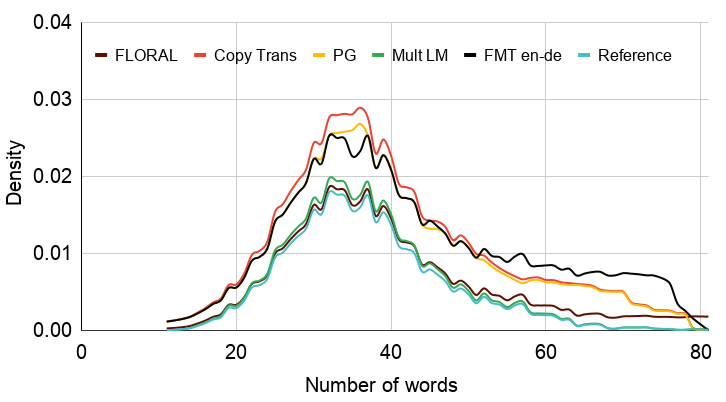}}\hspace{0.2cm}}
  \scalebox{0.88}{
  \subfigure[Density curve on AVIATE dataset.]{\includegraphics[scale=0.365]{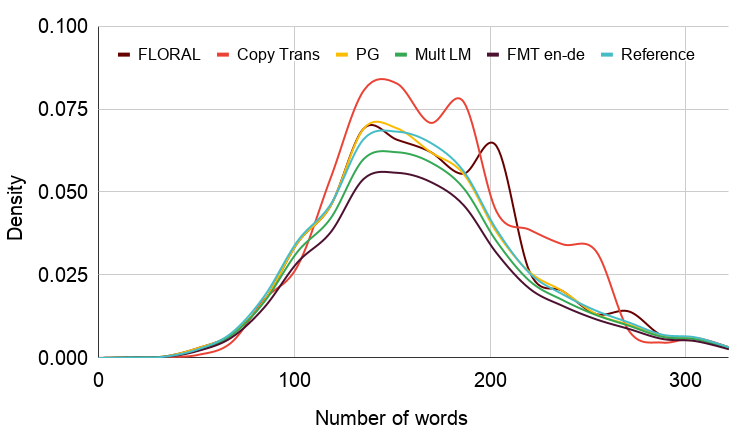}}}
  \vspace{-3mm}
  \caption{Word distribution of machine-generated summaries in comparison with the ground-truth summaries for different unimodal and multimodal systems on How2 and AVIATE datasets.}
  \label{fig:densitymap}
%   \vspace{-2mm}
\end{figure*}

\subsection{Qualitative Analysis}
In addition to ROUGE scores, we conduct a qualitative analysis by performing a human evaluation to understand the standard of the summary outputs. Following the abstractive summarization human annotation work of \citet{grusky2018newsroom}, the summaries were evaluated by five annotators\footnote{We employed five annotators who are experts in NLP, and their age ranges between 24-35 years.} to rate the generated summaries on a scale of $[1 - 5]$ on four parameters - informativeness, relevance, coherence, and fluency. For the evaluation, we randomly sampled $300$ videos from the test sets of How2 and AVIATE. Table \ref{tab:human} shows the average human evaluation scores for $4$ text-only, $1$ video-only and $3$ multimodal models. In general, we observe that PG has difficulty in summarizing articles with repetitive information and tends to assign a lower priority to less occurring important keywords. The extractive summarization systems sometimes pick sentences extraneous to the summary. For example, we notice some summaries generated by KLSumm starting with \textit{``Good afternoon everyone, I am $\cdot$''}, which is the very first line of the transcript. In contrast, the multimodal summarization models generate summaries with greater relevance and informativeness. Our proposed \name\ model obtains high scores on informativeness, relevance, and coherence on AVIATE, but sometimes seems to generate less fluent summaries. This fluency problem mostly stems from errors in ASR and OCR generated text. Some of these phenomena are illustrated with instances from AVIATE  in Table \ref{tab:output_examples}.

We also analyze the word distributions of the ground-truth summaries and different system-generated summaries. The density curves in Figure \ref{fig:densitymap} shows that for both How2 and AVIATE, the abstractive unimodal and multimodal summarization models generate summaries shorter than the ground-truth summary. The average length of summaries is highest for CopyTransformer. Interestingly, \name\ and PG generated summaries are similar in length. However, \name\ outperforms PG by a large margin, which illustrates that for improvements in ROUGE scores, an informative summary is more crucial than a lengthier summary.

\begin{table*}[hbtp]
  \centering
  \setlength{\extrarowheight}{12pt}
  \scalebox{0.5}{
  \begin{tabular}{p{2.7cm} | c | p{15.5cm}}
    \toprule
    \bf Model & \bf R-L & \bf Output \\
    \hline
    \hline
    \midrule
    Ground-truth   & - & \small We study the problem of semi-supervised question answering—utilizing unlabeled text to boost the performance of question answering models. We propose a novel training framework, the Generative Domain-Adaptive Nets. In this framework, we train a generative model to generate questions based on the unlabeled text, and combine model-generated questions with human-generated questions for training question answering models. We develop novel domain adaptation algorithms, based on to learn richer context-aware model on the insight and the human-generated data distribution. Experiments show that our proposed framework obtains substantial improvement from unlabeled text. \\ [0.05cm]

    \name\ (ours) & 31.47 & \small We study \textcolor{red}{the problem of semi-supervised} concepts unlabeled text to boost the \textcolor{blue}{performance of without without models}. We propose a novel training framework, \textcolor{red}{the Generative Domain-Adaptive Nets. In this framework, we train} a database model and generate concepts based on procedure on how partitioning in the levels between t, \textcolor{red}{and combine model-generated} \textcolor{blue}{attention with} human-generated \textcolor{red}{types for training associations} directed-generation models. We develop novel algorithms, based on model, to understand \textcolor{blue}{the discrepancy between} the generated data results \textcolor{red}{ and the human-generated data distribution. Experiments show} model obtains improvement from text. \\  [0.05cm] 
    MultT LM  & 28.73 & \small In learn paper are neural paper, we propose a new model and the paper, we \textcolor{blue}{study demonstrated the character score} We \textcolor{red}{investigate the problem answering utilizing unlabeled text} to understand trained models. we train a understanding model to generate understanding based on the text, and \textcolor{red}{combine trained understanding with human generated} understanding to training data trained we \textcolor{blue}{assume like, on a public unsupervised Subsequently}, \textcolor{red}{prediction trained to learn richer context-aware model on the insight} In trained facilitates smaller annotations previously reported understanding understanding on approach and questions \textcolor{blue}{along five perform two then} performance The the semantic best dataset and the proposed \\ [0.05cm]
    MultT En-De & 27.2 & \small We present the problem of supervised new \textcolor{red}{utilizing text of} new new models. We propose a problem model review a new model In this paper, we propose a simple model Experimental results\textcolor{red}{ prove the standard model achieves} method to then the proposed among review model to also performance based on the \textcolor{blue}{proposed CoNLL-2012 dataset}. \textcolor{red}{In this we train a new model to generate a} model and combine model. \textcolor{blue}{of machine morphology into more prediction on the drawback of the system classes}. \textcolor{blue}{using error-correcting codes collection.} We evaluate \textcolor{blue}{investigation of NMT and induced Experiments errors. shed light affect vs. scores} on a word its dataset with the proposed based \\ [0.05cm]
    CopyTrans.  & 27.06 & \small We propose a novel method for novel method for state-of-the-art \textcolor{red}{question answering} we show that the  method achieves state-of-the-art performance on the state-of-the-art performance of the proposed method on the systems \textcolor{blue}{over sequential text tasks or as independent, parallel tasks.} \textcolor{red}{In this framework model generate text} trained \textcolor{blue}{benefits of the new semantic high a image} and at them as \\ [0.05cm]
    HA  & 26.12 & \small We propose the novel model has the paper, we \textcolor{red}{propose a generative model} In both learning can can the benchmark model To capable to \textcolor{blue}{also performance of existing the pipelined of the model-generated model}: distribution for the \textcolor{red}{human-generated data distribution}. Experiments show that our proposed framework to the consistency of the \textcolor{blue}{art data distribution and event feedforward network}. The an unsupervised dependency framework performance and the human data distruction . \\ [0.05cm]
    PG  & 23.81   & \small In this paper we consider the problem of learning a deep neural network, \textcolor{red}{we propose a novel neural network architecture} based on novel neural network architecture that is \textcolor{blue}{trained end-to-end trainable convolutional neural network (CNN) architecture} \textcolor{red}{is able to train} \textcolor{blue}{a convolutional neural network architecture} that is capable of achieving state-of-the-art performance compared to state-of-the-art performance on three benchmark datasets. \\ [0.2cm]
   PG-MMR & 22.63 & \small We propose a novel method \textcolor{blue}{for object detection based on a novel object detection method} that uses a \textcolor{blue}{novel model based on the posterior distribution of the posterior distribution} over the parameters of the \textcolor{blue}{number of claims} and We show that  models can be used as well as compared to In this paper, we study the a algorithm that the proposed method can  outperforms the state-of-the-art methods on a large number of \\  [0.05cm]
    \bottomrule
    \hline
\end{tabular}}
  \caption{Comparison of ground-truth summary and outputs of $7$ different unimodal and multimodel abstractive text summarization systems - \name, MultT LM, MultT Encoder-Decoder, CopyTransformer, multimodal hierarchical attention (HA), Pointer Generator (PG) and Pointer Generator with MMR (PG-MMR) - arranged in the order of best to worst ROUGE-L scores in this table. \textcolor{red}{Red} highlighted text indicates a \textcolor{red}{positive} correlation of context w.r.t. ground-truth summary while \textcolor{blue}{blue} color represents a \textcolor{blue}{negative} correlation with ground-truth summary.}
  \label{tab:output_examples}
\end{table*}

\section{Conclusion}

In this paper, we explore the role of multimodality in abstractive text summarization. All the previous studies in this direction have used either images or short videos as the visual modality and generate one or two lines long summary, and thus, fail to perform on longer videos. Moreover, there exists no benchmark dataset for abstractive text summarization of medium and long videos. In this work, we introduce AVIATE, the first large-scale dataset for abstractive text summarization with videos of diverse duration, compiled from paper presentation videos in renowned academic conferences. We then propose \name, a Factorized Multimodal Transformer based decoder-only Language Model, which uses an increasing number of self-attentions to inherently capture inter-modal and intra-modal dynamics within the asynchronous multimodal sequences, without encountering difficulties during training even on relatively low-resource setups. To evaluate \name, we perform extensive experiments on How2 and AVIATE datasets and compare them against several unimodal and multimodal baselines. Overall, \name\ achieves superior performance over previously proposed models across two datasets. 

\section*{Acknowledgement} The work was partially supported by Ramanujan Fellowship (SERB) and the
Infosys centre for AI, IIIT Delhi, India.



\bibliographystyle{elsarticle-num-names.bst}

\bibliography{bibliography.bib}

\end{document}